\documentclass{article} 
\usepackage{iclr2025_conference,times}

\usepackage{amsmath,amsfonts,bm}

\def\eqref#1{equation~\ref{#1}}

\def\1{\bm{1}}

\DeclareMathAlphabet{\mathsfit}{\encodingdefault}{\sfdefault}{m}{sl}
\SetMathAlphabet{\mathsfit}{bold}{\encodingdefault}{\sfdefault}{bx}{n}

\usepackage{hyperref}
\usepackage{url}
\usepackage{graphicx}
\usepackage{overpic}

\usepackage[mathscr]{euscript}

\usepackage{booktabs}       

\usepackage[capitalize,noabbrev]{cleveref}

\usepackage{booktabs} 
\usepackage{caption, copyrightbox}

\newcommand{\grasp}{\revised{GGSD}}

\newcommand{\red}[1]{{#1}}

\newcommand{\revised}[1]{{#1}}

\DeclareCaptionLabelFormat{andtable}{#1~#2  \&  \tablename~\thetable}

\title{Generating  Graphs via Spectral Diffusion}

\iclrfinalcopy

\author{Giorgia Minello \\
Ca' Foscari University\\
\texttt{giorgia.minello@unive.it} \\
\And
Alessandro Bicciato \\
Ca' Foscari University\\
\texttt{alessandro.bicciato@unive.it} \\
\And
Luca Rossi \\
The Hong Kong Polytechnic University\\
\texttt{luca.rossi@polyu.edu.hk} \\
\And
Andrea Torsello \\
Ca' Foscari University\\
\texttt{andrea.torsello@unive.it} \\
\And
Luca Cosmo \\
Ca' Foscari University\\
\texttt{luca.cosmo@unive.it} \\
}

\renewcommand{\lhead}[1]{}
\iclrfinalcopy 

\begin{document}
\maketitle

\begin{abstract}
In this paper, we present \grasp, a novel graph generative model based on 1) the spectral decomposition of the graph Laplacian matrix and 2) a diffusion process. Specifically, we propose to use a denoising model to sample eigenvectors and eigenvalues from which we can reconstruct the graph Laplacian and adjacency matrix. Using the Laplacian spectrum allows us to naturally capture the structural characteristics of the graph and work directly in the node space while avoiding the quadratic complexity bottleneck that limits the applicability of other diffusion-based methods. This, in turn, is accomplished by truncating the spectrum, which, as we show in our experiments, results in a faster yet accurate generative process, and by designing a novel transformer-based architecture linear in the number of nodes. Our permutation invariant model can also handle node features by concatenating them to the eigenvectors of each node. An extensive set of experiments on both synthetic and real-world graphs demonstrates the strengths of our model against state-of-the-art alternatives.
\end{abstract}

\section{Introduction}\label{sec:intro}
Generating realistic graphs by learning from a distribution of real-world graphs has gained increasing attention from researchers in many fields due to its wide range of applications. For instance, synthetic graph generation plays a crucial role in drug design~\cite{gomez2018automatic,li2018multi,you2018graph} as well as in network science~\cite{watts1998collective,leskovec2010kronecker,albert2002statistical}.

\revised{Seminal} graph generation approaches date back to the 1960s and rely on simple stochastic processes, limiting their ability to capture complex dependencies seen in real-world networks. For example, the Barabási-Albert~\cite{albert2002statistical} and Kronecker~\cite{leskovec2010kronecker} graph models are specifically designed to generate graphs belonging to specific families and lack the ability to learn directly from observed data. While these models may excel in capturing a set of predefined properties, they are often unable to represent a wider range of aspects observed in real-world graphs. In addition, in several domains, network properties are largely unknown, which further limits the applicability of these techniques. For instance, the Barabási-Albert model~\cite{albert2002statistical} allows to create graphs that exhibit the scale-free nature found in empirical degree distributions, however it is unable to capture other facets of real-world graphs, {\em e.g.}, community structure. \revised{While a flurry of new models attempting to address these shortcomings have been introduced by the network science community (see~\cite{drobyshevskiy2019random} for a recent review), these methods often lack the ability to learn to mimic the characteristics of a given dataset. This in turn limits the expressivity and fidelity of generated graphs and thus the range of possible applications of graph generative models.}

\revised{In this paper, we introduce a new model for Generating Graphs via Spectral Diffusion (\grasp). }
The ideas underpinning our approach are 1) to represent the graph using the eigendecomposition of its Laplacian matrix and 2) to use a diffusion-based approach to learn to sample sets of eigenvalues and eigenvectors from which a graph adjacency matrix can be reconstructed. Doing this allows us to work directly in the space of nodes while overcoming the computational bottleneck (quadratic in the number of graph nodes) of other methods that follow a similar approach~\cite{vignac2022digress}. 
%
By limiting the number of eigenvalues and eigenvectors used to reconstruct the graph adjacency matrix, we reduce the complexity of the iterative denoising process to be linear with respect to the number of nodes while, at the same time, having a representation tablet to encapsulate graph structural characteristics. 
Moreover, unlike other models conditioned on spectral representations~\cite{martinkus2022spectre}, our model also allows us to robustly condition the generation of new graphs on desired spectral properties (subsets of eigenvalues and/or eigenvectors) at inference time.

The remainder of this paper is structured as follows. Section~\ref{sec:related} reviews the related work, while Sections~\ref{sec:diffusion} introduces the necessary background on denoising diffusion models. We introduce our graph generative model in Section~\ref{sec:model}, and we present the experimental evaluation against state-of-the-art alternatives in Section~\ref{sec:experiments}. Finally, Section~\ref{sec:conclusion} concludes the paper.

\section{Related Work}\label{sec:related}

In contrast to the image and text domains, where the development of generative models is well understood and established, graphs introduce a series of additional challenges.

The first issue is the non-uniqueness of graph representations, {\em i.e.}, if a graph contains $n$ nodes, there exist up to $n!$ \revised{possibly} distinct adjacency matrices that serve as equivalent representations of the same graph, \revised{since there is no reason to prefer a particular node order}. Ideally, a generative model should assign equal probability to each of these $n!$ adjacency matrices. Another crux lies in the size of the output space, which is quadratic in the number of nodes, and that quickly becomes a bottleneck when dealing with large graphs. Graph generative models should also be able to consider the existence of dependencies and relationships between nodes and edges, rather than treating them as independent, {\em e.g.}, in social networks the likelihood of two nodes being connected is often higher when they have common neighbors. Finally, standard machine learning techniques designed for continuously differentiable objective functions are unsuitable to be directly applied to discrete graph structures~\cite{guo2022systematic}. 

\revised{Seminal} graph generative model approaches seek to address these problems, yet focus only on the generation of graphs displaying a limited set of structural characteristics. These initial methods rely on identifying common characteristics in real-world graphs, such as degree distribution, graph diameter, and clustering coefficient~\cite{faloutsos2008graph}, and then generate synthetic graphs through the application of a set of heuristic rules~\cite{leskovec2010kronecker,leskovec2007scalable, erdHos1960evolution, albert2002statistical}. Although these models can produce synthetic graphs with the given desired features, they are limited in their ability to generate node features as well as novel structural patterns.

A breakthrough in this field has been marked by the recent progress in deep learning models such as Variational Auto Encoders (VAEs)~\cite{kingma2013auto}, Recurrent Neural Networks (RNNs)~\cite{zaremba2014recurrent} and Generative Adversarial Networks (GANs)~\cite{goodfellow2014generative}. In this context, we encounter models commonly referred to as \textit{general-purpose deep graph generative models}, such as GraphRNN~\cite{you2018graphrnn} and GRAN~\cite{liao2019efficient}, which exploit deep architectures to learn the graphs distribution. Even though they represent a step forward in the field of generative graph models, most of them are limited by exclusively focusing on the graphs structure. Further, approaches of this type adopt evaluation metrics based only on graph statistics, like degree distribution or clustering coefficients, and thus overlook or omit the assessment of the generated node features.

Node and edge features are instead considered in a number of methods developed specifically for the generation of molecules, indeed one of the most promising application scenario for modern graph generation approaches. Models falling in this domain, referred to as \textit{molecule graph generative models}, exploit deep architectures such as GAN in~\cite{de2018molgan} or RNN in~\cite{popova2019molecularrnn} as well as other generation strategies ({\em e.g.}, graph normalizing flows~\cite{luo2021graphdf}) or the combination of different approaches. For instance, \cite{shi2020graphaf} combines the advantages of both autoregressive and flow-based methods.

Nevertheless, there are other deep learning approaches beyond molecule graph generative models that are capable of generating graphs with node and edge features - even though the evaluation itself is often still based on a molecule generation task. For instance~\cite{simonovsky2018graphvae} and~\cite{grover2019graphite} propose general deep generative models for graphs based on variational autoencoders. The main drawback of these architectures is that they are specialized and limited to small-scale graphs with low-dimensional feature space~\cite{yoon2023graph}.

Another category of graph generative models takes cues from the score-based generative modeling work of~\cite{song2019generative} to define diffusion models for graphs. For instance, in~\cite{huang2022graphgdp} the authors propose a forward diffusion process, specifically a continuous-time generative diffusion process for permutation invariant graph generation. Similarly, \cite{niu2020permutation} introduce a different diffusion model named Edge-wise Dense Prediction Graph Neural Network (EDP-GNN), which uses Gaussian noise and uses thresholding to address the issue of generating a discrete valued adjacency matrix from continuous values. Crucially, the proposed method cannot fully capture node-edge dependencies. A similar score-based generative model for graphs, where both node features and adjacency matrix are created, is presented in~\cite{jo2022score}. Finally, \cite{vignac2022digress} suggest an alternative approach where a discrete diffusion process is used to generate graphs with discrete node and edge features. This is similar to~\cite{haefeli2022diffusion}, however the latter can only be applied to unattributed graphs.

More recently, \cite{martinkus2022spectre} with their SPECTRE network and~\cite{luo2023fast} take a different approach by considering the graph spectra, thus leveraging the inherent ability of the low frequency portion of the spectrum to capture global structural characteristics of the corresponding graph. \revised{Although similar to our method, SPECTRE focuses on generating an adjacency matrix conditioned on a set of eigenvectors, which may or may not have been generated themselves. Our method instead only generates eigenpairs from which the adjacency matrix is recovered. As a result, unlike  SPECTRE, our method is capable of generating graphs that respect a set of given spectral properties (see Subsection~\ref{subsec:conditioning}). GSDM~\cite{luo2023fast}, on the other hand, proposes to reduce the complexity by performing diffusion just on the eigenvalues and optionally on the node features, while the eigenvectors used to reconstruct the final adjacency matrix are uniformly sampled from the training set.} DiGress~\cite{vignac2022digress} is also closely related to our model, however its complexity is quadratic in the number of nodes of the graph, making it unsuitable to work on large graphs.

\revised{Our approach is also related to recently introduced latent graph diffusion models, which employ an autoencoder architecture to map nodes and edges of a graph to latent space where the diffusion process takes place~\cite{yang2024graphusion,zhou2024unifying}. While these approaches aim to learn a low-dimensional embedding of the graph nodes, we rely instead on the well-established concept of spectral embedding~\cite{luo2003spectral}, with eigenvectors providing a low-dimensional embedding of the graph nodes and eigenvalues capturing global structure information.}
 
\section{Denoising Diffusion Models}\label{sec:diffusion}

Denoising Diffusion Probabilistic Models (DDPMs) are a class of generative models inspired by considerations from non-equilibrium thermodynamics. In particular, diffusion models in deep learning were first introduced in~\cite{sohl2015deep} yet popularized only in 2020~\cite{ho2020denoising}. They operate by iteratively introducing noise to an input signal and then learning to denoise it thus generating new samples from the corrupted signals.  
Specifically, the idea is to destroy the structure in a data distribution through an iterative forward diffusion process (noising) and then learn a reverse diffusion process (denoising). This reverse process restores structure in the data, thus yielding a tractable generative model of the data.

Given a data point sampled from a real but unknown data distribution $\mathbf{x_0} \sim q(\mathbf{x})$, we define a forward noising process $q$ producing a sequence of noisy samples $\mathbf{x}_1, \ldots, \mathbf{x}_T$ as a Markov Chain given by
$q\left(\mathbf{x}_1, \ldots, \mathbf{x}_T \mid \mathbf{x}_0\right) =\prod_{t=1}^T q\left(\mathbf{x}_t \mid \mathbf{x}_{t-1}\right)$, with the diffusion kernel defined as:
\begin{equation}
\begin{aligned}
q\left(\mathbf{x}_t \mid \mathbf{x}_{t-1}\right) = \mathcal{N}\left(\mathbf{x}_t ; \sqrt{1-\beta_t} \mathbf{x}_{t-1}, \beta_t I\right).
\end{aligned}\label{eq:diffusion_kernel}
\end{equation}
Note that, if we define  $\bar{\alpha}_t=\prod_{s=1}^t\left(1-\beta_s\right)$, we can reformulate Eq.~\ref{eq:diffusion_kernel} as a single step
\begin{equation}
q(\mathbf{x}_t \mid \mathbf{x}_0)=\mathcal{N}\left(\mathbf{x}_t ; \sqrt{\bar{\alpha}_t}  \mathbf{x}_0,\left(1-\bar{\alpha}_t\right) \mathbf{I}\right)\,.
\label{eq:forward_diffusion_xt_x0}
\end{equation} 


In the reverse diffusion process, the goal is to recreate the true sample from a Gaussian noise input $\mathbf{x}_T\sim \mathcal{N}(\mathbf{0},\mathbf{I})$ by sampling from $q\left(\mathbf{x}_{t-1} \mid \mathbf{x}_t\right)$, the true denoising distribution. 
In order to run the reverse diffusion process, we need to learn a model $p_\theta$, often referred to as \textit{score model}, to approximate these conditional probabilities.
As~\cite{feller49} showed, in the case of Gaussian distributions the diffusion process reversal has the same functional form of the forward process. From this it follows that the reverse diffusion process kernel can be defined as
 \begin{equation}
p_\theta(\mathbf{x}_{t-1} \mid \mathbf{x}_t)=\mathcal{N}(\mathbf{x}_{t-1} ; \boldsymbol{\mu}_\theta(\mathbf{x}_t, t), \boldsymbol{\Sigma}_\theta(\mathbf{x}_t, t))\,,
\end{equation}

where $\theta$ are the parameters of the reverse diffusion kernel at each time step, which can be learned using a neural network. If we fix the variance to a constant $\beta_t$ (\textit{i.e.} $\boldsymbol{\Sigma}_\theta(\mathbf{x}_t, t) = \beta_t I$), we only need  to learn the distance between the means of two Gaussian distributions, {\em i.e.}, between the noise added in the forward process and the noise predicted by the model. This leads to a variational lower bound loss expressed in terms of the Kullback–Leibler (KL) divergence between the posterior of the forward process and the parameterized reverse diffusion process.

\begin{figure*}[t!]
\centering
\includegraphics[width=0.9\textwidth]{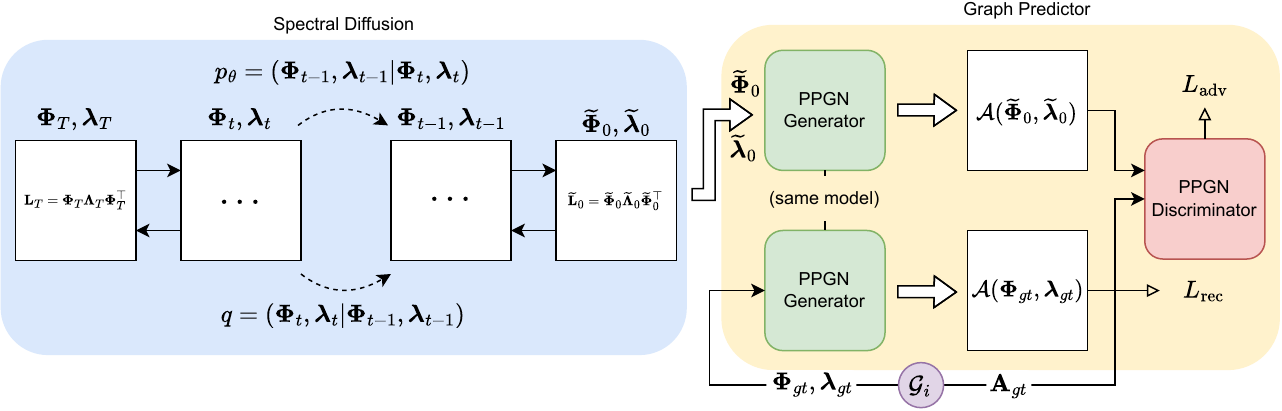}
\caption{\grasp~pipeline. During the spectral diffusion process (left) the neural network is trained to predict the denoising steps for the eigenvectors $\phi$ and eigenvalues $\lambda$ of the graph Laplacian. The second stage of our method is the graph predictor (right), where we train a Provably Powerful Graph Network (PPGN) \cite{NEURIPS2019_bb04af0f} (similar to what was done in SPECTRE \cite{martinkus2022spectre}). Given the eigenvalues and eigenvectors generated, it predicts the adjacency matrix.}   
\label{fig:pipeline}
\end{figure*}

\section{Our Method}\label{sec:model}

Consider an undirected unweighted graph $\mathcal{G} = (\mathcal{V},\mathcal{E})$, where $\mathcal{V}$ is the set of $n$ nodes connected by the edge set $\mathcal{E}$. \revised{Recall that for a graph with adjacency matrix $\mathbf{A}$, the graph Laplacian $\mathbf{L}$ is defined as $\mathbf{L} = \mathbf{D} - \mathbf{A}$, where $\mathbf{D}$ is the diagonal degree matrix. Finally, let $\mathbf{\Phi}$ and $\mathbf{\Lambda}$ be the orthonormal matrix of eigenvectors (as columns) and the diagonal matrix of eigenvalues given by the eigendecomposition $\mathbf{L} = \mathbf{\Phi} \mathbf{\Lambda} \mathbf{\Phi}^{\top}$, respectively. In the following sections, we use $\boldsymbol{\lambda}$ to denote the vector of eigenvalues of the graph Laplacian.}



The fundamental intuition underpinning our model is that we can represent the graph connectivity with (possibly a subset of) the eigenvectors $\mathbf{\Phi}$ and the corresponding eigenvalues $\boldsymbol{\lambda}$ of the graph Laplacian. The connection between the spectrum of the graph Laplacian and the structural properties of the underlying graph is well known and studied. For example, it is well established that the low frequency portion of the spectrum captures the global structural characteristic of the graph, while the high frequencies are essential in the reconstruction of local connectivity patterns~\cite{chung1997spectral}.

\revised{Fig.~\ref{fig:pipeline} shows an overview of the proposed pipeline. This is made of two main components, namely 1) a spectral diffusion process that generates a set of eigenvalues and eigenvectors from which an approximation of the Laplacian matrix can be reconstructed and 2) a graph predictor, which outputs a binary adjacency matrix from the (noisy) Laplacian reconstruction. The two components are discussed in detail in the following subsections.}

\subsection{Spectral Diffusion}
Moving from the Laplacian to its spectrum reduces the double row-, column-covariance  with respect to node permutations of the Laplacian matrix, to a single covariance over the rows of the eigenvector matrix. To address this covariance, we represent the eigenvector matrix as a series of spectral embedding of the nodes, {\em i.e.}, we interpret the $i$-th component of the $j$-th eigenvector as the $j$-th component of the $i$-th node embedding, or alternatively, we see the rows of $\mathbf{\Phi}$ as vectors.
To this we add the eigenvalues $\boldsymbol{\lambda}$ as a global graph descriptor. We can then define the \revised{reverse} diffusion step as 
\begin{equation}    
p_\theta(\mathbf{\Phi}_{t-1},\boldsymbol{\lambda}_{t-1} | \mathbf{\Phi}_t,\boldsymbol{\lambda}_t) =  
\mathcal{N}\left(\{\mathbf{\Phi}_{t-1},\boldsymbol{\lambda}_{t-1}\} ; \boldsymbol{\mu}_\theta\left(\mathbf{\Phi}_t,\boldsymbol{\lambda}_t, t\right), \sigma_t^2 \mathbf{I}\right) \,
\label{eq:denoise_step}
\end{equation}
where the normal distribution is over the product set of the spectral embeddings and the global spectral descriptor.

\begin{figure}[t!]
\centering
\includegraphics[width=0.7\columnwidth]{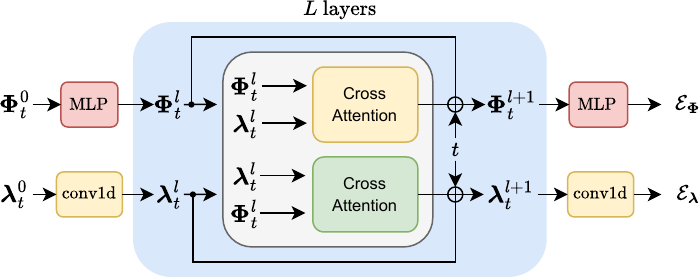}
\caption{The score model takes as input the noisy eigenvector matrix and eigenvalues at time $t$ and predicts the noise of the data to be used in the denoising step. The $k$ node feature eigenvectors $\mathbf{\Phi}_t^0$ are projected through an MLP to a $d$ dimensional space. The sequence of $k$ eigenvalues is given as input to a 1D convolutional layer, which outputs $d$ features for each eigenvalue. Both eigenvectors and eigenvalues go through a series of $L$ layers composed of two multi-head cross-attention blocks, one updating the eigenvectors conditioned by the eigenvalues and one updating the eigenvalues conditioned on the eigenvectors. After each layer, we apply a residual block $\oplus$, which adds to the layer input the updated values scaled and shifted by time-dependent factors. Finally, $\mathbf{\Phi}_t^L$ and $\boldsymbol{\lambda}_t^L$ are projected to a $k$ dimensional space through an MLP and a 1D convolution.}
\label{fig:score_model}
\end{figure}

Following DDPM \cite{ho2020denoising}, we train a neural network to predict the denoising step. We design the backbone of our spectral diffusion process of Eq.~\ref{eq:denoise_step} as a neural network composed of a sequence of layers containing a pair of multi-head attention blocks~\cite{NIPS2017_3f5ee243}, one operating on the eigenvectors conditioned on the eigenvalues and one operating on the eigenvalues conditioned by the eigenvectors. This choice allows us to achieve both a node permutation invariant model and to let the eigenvectors and eigenvalues condition each other on the prediction of the denoising step. Moreover, this model can easily handle node features $\mathbf{X}$ by simply concatenating them to the eigenvectors of each node. Fig.~\ref{fig:score_model} shows the overall structure of the proposed neural network. Note that the \texttt{conv1d} layer is applied to the eigenvalues, which are invariant with respect to node permutations, while the diffusion process itself acts in an invariant way on node embeddings, which are permutationally covariant. As a consequence, the whole process is permutationally invariant.

\revised{Crucially, our model allows us to} reduce the memory footprint of the diffusion component from $O(n^2)$ to $O(kn)$, where $n$ denotes the number of graph nodes, by fixing the maximum number of eigenvectors to $k$, resulting in a faster generative process. For this reason, for larger graphs, we perform diffusion on a subset of $k$ eigenvectors. In this case, after the denoising diffusion process, we obtain a subset $\tilde{\mathbf{\Phi}}$ of columns of $\mathbf{\Phi}$ with the corresponding subset of eigenvalues $\tilde{\boldsymbol{\lambda}}$, which allows for an approximated reconstruction of $\tilde{\mathbf{L}} = \tilde{\mathbf{\Phi}}\tilde{\mathbf{\Lambda}}\tilde{\mathbf{\Phi}} \approx \mathbf{L}$, from which the adjacency matrix can be inferred. \revised{In addition, since the eigenvector associated with the null eigenvalue does not contribute to the reconstruction of the Laplacian matrix, we can safely ignore it in the generation process.}

Note that the optimal number of eigenvalues/vectors ($k$) to use is not fixed. We determined its range through preliminary analyses on the SBM and Planar datasets (see \Cref{sec:num_eigenvectors}), and in general, we treat it as a hyper-parameter of the model. 
Interestingly, our experiments reveal that the eigenvectors corresponding to the smallest eigenvalues (lowest frequencies) do not consistently offer more information about connectivity or result in better reconstructions of the original adjacency matrix.



\subsection{Graph Predictor}
The main drawback of considering a subset of the eigenvectors is the introduction of noise on the reconstructed adjacency matrix. We adopt a strategy similar to the one proposed in SPECTRE~\cite{martinkus2022spectre} to predict a binary adjacency matrix starting from a noisy reconstruction. We train a {\it l}-layer Provably Powerful Graph Network (PPGN)~\cite{NEURIPS2019_bb04af0f}, which takes as input the generated eigenvectors $\tilde{\mathbf{\Phi}}$ scaled by the square root of the eigenvalues $\tilde{\boldsymbol{\lambda}}$ as node features as well as the noisy adjacency matrix $\tilde{\mathbf{A}}=\tilde{\mathbf{D}}-\tilde{\mathbf{L}}$ and predicts the binary adjacency matrix
\begin{equation}
\mathcal{A}(\tilde{\mathbf{\Phi}},\tilde{\boldsymbol{\lambda}}) = \sigma(\textrm{PPGN}_{\textrm{{\it l}}}(\tilde{\mathbf{A}}, \tilde{\mathbf{\Phi}}\tilde{\mathbf{\Lambda}}^{\frac{1}{2}})),
\end{equation}
where $\sigma$ is a sigmoid activation function and $\textrm{PPGN}_{\textrm{{\it l}}}$ is a sequence of {\it } PPGN layers.
We train this network with two losses: 1) a reconstruction loss on the eigenvectors $\tilde{\mathbf{\Phi}}_{gt}$ and eigenvalues $\tilde{\boldsymbol{\lambda}}_{gt}$ of the reduced spectrum of the training graphs with adjacency matrix $\mathbf{A}_{gt}$ and 2) an adversarial loss on the generated adjacency matrix $\mathcal{A}(\tilde{\mathbf{\Phi}},\tilde{\boldsymbol{\lambda}})$, {\em i.e.},
\begin{equation}
L_{\textrm{rec}} = \textrm{BCE}(\mathbf{A}_{gt},\mathcal{A}(\tilde{\mathbf{\Phi}}_{gt},\tilde{\boldsymbol{\lambda}}_{gt}))
~\mbox{ and }~
L_{\textrm{adv}} = \log(\mathcal{D}(\mathbf{A}_{gt}))+\log(1-\mathcal{D}(\mathcal{A}(\tilde{\mathbf{\Phi}},\tilde{\boldsymbol{\lambda}})))\,,
\end{equation}
where BCE is the standard binary cross entropy loss and $\mathcal{D}$ is a discriminator network composed of a sequence of PPGN layers followed by a global pooling for the final graph-level classification. Note that, unlike in SPECTRE~\cite{martinkus2022spectre}, this refining step is not generative, meaning that the output is deterministic and depends solely on the input eigenvectors/values.

\section{Experimental Evaluation}\label{sec:experiments}

\paragraph{Datasets.}
We compare the performance of our model against that of state-of-the-art alternatives on both synthetic and real-world datasets. As commonly done in the literature, we use three synthetic datasets and two real-world datasets. The synthetic datasets we consider are (i) Community-small ($12 \leq |V| \leq 20$), (ii) Planar ($|V|=64$), and (iii) Stochastic Block Model (SBM) (2-5 communities and 20-40 nodes per community). The real-world datasets are both from the molecular domain, namely (i) Proteins (100-500 nodes)~\cite{dobson2003distinguishing} and (ii) QM9 (9 nodes)~\cite{ruddigkeit2012enumeration,
ramakrishnan2014quantum}. Detailed descriptions of all datasets can be found in Appendix~\ref{sec:dataset}.

\vspace{-3mm}\paragraph{Evaluation Metrics.}
We assess the ability of the models to generate graphs with structural characteristics close to those of the training graphs by following the methodology outlined in~\cite{liao2019efficient}, which aims to address the difficulties of measuring likelihoods when evaluating autoregressive graph generative models reliant on orderings. In particular, we adopt the approach proposed by~\cite{you2018graphrnn} and~\cite{li2018learning} and used by many others~\cite{krawczuk2020gg} as well. Our evaluation centers on contrasting the distributions of graph statistics between the generated and actual graphs. Specifically, we consider the following key graph statistics: 1) degree distribution (Deg.), 2) clustering coefficient (Clus.), 3) eigenvalues of the normalized graph Laplacian (Spec.), and 4)  the occurrence frequency of all 4-node orbits (Orb.).
Moreover, for QM9, we follow the literature and evaluate the quality of the generated graphs by computing the validity of the generated molecules, their uniqueness, and their novelty w.r.t. to the molecules in the training set and vicerversa~\cite{samanta2020nevae,guo2022systematic}. 
Detailed information on the evaluation metrics used can be found in Appendix~\ref{sec:metrics}.



\begin{table}[!t]\centering
\caption{ Comparison with other graph generative models using MMD metrics (the smaller, the better) on synthetic datasets.}\label{tab:synth}
\scriptsize
\setlength{\tabcolsep}{4.5pt}
\begin{tabular}{lcccc|cccc|cccc}\toprule
&\multicolumn{4}{c}{\textbf{Community-Small}} &\multicolumn{4}{c}{\textbf{Planar}} &\multicolumn{4}{c}{\textbf{Stochastic Block Model (SBM)}} \\
\cmidrule{2-13}
&Deg. $\downarrow$ &Clus. $\downarrow$ &Spect. $\downarrow$ &Orb. $\downarrow$ &Deg. $\downarrow$ &Clus. $\downarrow$ &Spect. $\downarrow$ &Orb. $\downarrow$ &Deg. $\downarrow$ &Clus. $\downarrow$ &Spect. $\downarrow$ &Orb. $\downarrow$ \\
\midrule
\midrule
GraphRNN &0.0271 &0.1072 &0.0520 &0.1469 &0.0096 &0.2985 &0.0389 &1.4022 &0.0178 &0.0151 &0.0104 &0.0351 \\
GRAN &\textbf{0.0013} &{0.0843} &{0.0282} &{0.0201} &0.0202 &0.2985 &0.0248 &0.1964 &0.0135 &0.0149 &\underline{0.0034} &0.0352 \\
DiGress &0.0096 &0.1035 &0.0506 &0.0372 &\textbf{0.0005} &\textbf{0.0178} &\textbf{0.0020} & 0.0115 &0.0166 &0.0246 &0.0064 &0.1327 \\
\revised{GSDM} &0.0099 &\textbf{0.0446} &\textbf{0.0131} &\underline{0.0155} &0.0220 &\underline{0.0222} &\underline{0.0096} &0.0371 &0.2295 &0.2280 &0.1578 &0.2876 \\
GDSS &0.0107 &0.1060 &0.0450 &0.0356 &0.0701 &0.3025 &0.0403 &1.0345 &0.2658 &0.0442 &0.0551 &0.2780 \\
SPECTRE &0.0079 &0.1067 &0.0460 &0.0250 &0.0008 &0.0859 &0.0147 &\underline{0.0058} &\underline{0.0044} &\underline{0.0118} &\textbf{0.0015} &\textbf{0.0140} \\
\midrule
{\grasp} &\underline{0.0016} &\underline{0.0590} &\underline{0.0153} &\textbf{0.0142} &\underline{0.0007} &{0.1881} & 0.0125 &\textbf{0.0047} &\textbf{0.0005} &\textbf{0.0115} &0.0045 &\underline{0.0289} \\
\bottomrule
\end{tabular}
\end{table}

\vspace{-3mm}\paragraph{Baselines.}
We evaluate the effectiveness of our model by comparing it against a number of well-known graph generative models as well as some recently developed deep graph generative models. In particular, we consider GraphRNN~\cite{you2018graphrnn}, GRAN~\cite{liao2019efficient}, DiGress~\cite{vignac2022digress}, SPECTRE~\cite{martinkus2022spectre}, GDSS~\cite{jo2022score} and \revised{GSDM}~\cite{luo2023fast}. For the molecule generation task we also include GraphVAE~\cite{simonovsky2018graphvae}.

\vspace{-3mm}\paragraph{Experimental Setup.} To maximize the robustness of the experimental results, we follow a slightly different experimental setup compared to previous works. Specifically, for the synthetic datasets, we decided to create a larger set of test graphs: 200 graphs for Planar and SBM, and 100 graphs for community-small. Accordingly, we let each model generate an equivalent number of graphs (200 for Planar and SBM, 100 for community-small) to compute the MMD measures. Due to the limited number of graphs in the Proteins dataset (see Appendix~\ref{sec:dataset}), we also followed a different and more robust protocol to evaluate the generated graphs on this dataset. Rather than utilizing a single subset of the dataset as a test set, we created 10 folds (identical for each method) allowing us to report the average of each metric ($\pm$ standard deviation) over the 10 folds. Further detailed information on the model settings and training setup, both for our model and the baselines, is provided in Appendix~\ref{sec:modelsettings}.

\subsection{Evaluating the Generated Graphs}

\paragraph{Synthetic Datasets}
\phantomsection
\label{par:synth_res}

The experimental results for community-small, SBM, and Planar are shown in \Cref{tab:synth}.
In this table, we report the MMD metrics for the graph statistics, where the smaller the statistics, the better. The results of our method (\grasp) are chosen from those obtained using either the lower or upper range of eigenvalues. Specifically, for the Planar, the results refer to the 16 smallest eigenvalues, whereas for the community-small dataset and SBM we used the largest ones, 8 and 32 respectively. The best performance is highlighted in bold, while the second-best value is underlined. Overall, our model consistently achieves the best or second-best results across all datasets, with the exception of \textit{Clus.} in Planar and \textit{Spect.} in SBM and Planar. \revised{We posit that the lower performance on planar graphs may be related to the behaviour of the eigenvectors of this type of graphs. Note in fact that there is no clear class structure in this dataset but rather the graphs are related by the (hard) property of planarity. Indeed, graphs with similar spectra can lie on opposite sides of the discrimination boundary, {\em i.e.}, between planar and non-planar graphs. As such, the addition or removal of an edge connecting local substructures can easily break the planarity of the graph without significantly affecting its spectral representation. Finally, for  Planar and SBM, we also evaluated the quality of the generated graphs in terms of validity, uniqueness, and novelty in \Cref{sec:vun}, showing the ability of our method to generate graphs that are at the same time valid, unique, and novel.}

\vspace{-3mm}\paragraph{Real-world Datasets}
Results for real-world dataset generations on Proteins and QM9 are reported in \Cref{tab:proteins_qm9}, left and right, respectively. Also in this series of experiments, we  achieve good performance. Again, we selected the best results from either the highest or lowest frequencies. Notably, for the Proteins dataset, we utilized the 16 smallest eigenvalues, while for QM9, we used the entire spectrum.
The Proteins dataset is especially challenging due to the size of the graphs, which can reach 500 nodes. For this dataset, \grasp~ranks as the best in all metrics \Cref{tab:proteins_qm9} (left).
QM9, on the other hand, is composed by small graphs of up to 9 nodes, with both node and edge features. In \Cref{tab:proteins_qm9} (right), it is important to note that the last column is of particular interest as it summarizes all values, where we emerge as the second top performer.   

\begin{table*}
\caption{Comparison on real datasets with other graph generative models. \textbf{Left}: Proteins, using MMD metrics (the lower, the better) and mean $\pm$ standard deviation (over 10 folds). \textbf{Right:} QM9, based on (V)validity, (U)niqueness, and (N)ovelty metrics (the higher, the better). 
Results denoted by $^*$ and $^\dagger$ are taken from   \cite{martinkus2022spectre} and \cite{vignac2022digress} respectively.\label{tab:proteins_qm9}}
\scriptsize
\setlength{\tabcolsep}{3.5pt}
\centering 
\begin{tabular}[t]{@{}rcccccc@{}}
\toprule   & 
 \multicolumn{4}{c}{\textbf{Proteins}}\\ 
\cmidrule{2-5}  
& Deg. $\downarrow$ & Clus. $\downarrow$ & Spect $\downarrow$ & Orbit $\downarrow$  \\
\midrule 
\midrule 
GraphRNN &   \underline{0.0065$\pm$0.0011} & 0.1658$\pm$0.0088  & 0.0170$\pm$0.0009 & 0.8142$\pm$0.0273 \\

GRAN &  0.0569$\pm$0.0056  &  0.1622$\pm$0.0092  &  0.0146$\pm$0.0007&0.3430$\pm$0.0363\\

\revised{GSDM} & 0.3792$\pm$0.0041 & 0.4653$\pm$0.0089 & 0.3024$\pm$0.0032 & 0.9589$\pm$0.0285 \\

GDSS & 0.0653$\pm$0.0063 & 0.4160$\pm$0.0089 & 0.0706$\pm$0.0021 & 0.8168$\pm$0.0118 \\

SPECTRE &   0.0082$\pm$0.0021  &  \underline{0.0988$\pm$0.0071}  &  \underline{0.0066$\pm$0.0004}&\underline{0.0328}$\pm$0.0039\\

\midrule
\grasp & \textbf{0.0014$\pm$0.0003} &  \textbf{0.0856$\pm$0.0066}  &  \textbf{0.0059$\pm$0.0007} &\textbf{0.0296$\pm$0.0066}\\
 
\bottomrule 
\end{tabular} 
\hfill
\begin{tabular}[t]{@{}rccc@{}}
\toprule   & 
\multicolumn{3}{c}{\textbf{QM9} }       \\ 
\cmidrule{2-4}  
& Val.$\uparrow$&V.\& U.$\uparrow$&  V.\& U.\& N.$\uparrow$ \\
\midrule 
\midrule

GraphVAE$^*$ & 0.5570 & 0.4200 & 0.2610 \\

DiGress$^\dagger$ & 0.9900 & 0.9523 & {0.3180}\\

GDSS$\;\,$ & 0.8335 & 0.8281 & \underline{0.7257} \\

SPECTRE$^*$ & 0.8730 & 0.3120 & 0.2910\\
\midrule
{\grasp}$\;\,$ & 0.966 & 0.864 & \textbf{0.847} \\
\bottomrule 
\end{tabular} 
\end{table*}






 




\subsection{Graph Predictor Ablation} \label{sec:ablation}
Given that the ``Graph Predictor" is trained using a discriminative loss, one may think that it is this component that is doing all the ``heavy lifting" of the graph generation task, while the ``Spectral Diffusion" may be just producing noisy data.

To assess that this is not the case, we designed an experiment training our model to predict all the eigenvectors and eigenvalues on the community-small dataset. The small size of the graphs in this datasets allows us to generate all the eigenvectors $\mathbf{\Phi}$ and eigenvalues $\boldsymbol{\lambda}$ and reconstruct the exact Laplacian $\mathbf{L}=\mathbf{\Phi}\mathbf{\Lambda}\mathbf{\Phi}^{\top}$ and the (almost) binary adjacency matrix $\mathbf{A}=\mathbf{D}-\mathbf{L}$, where $\textbf{D}$ is the diagonal of the Laplacian. 
\red{To obtain a binary adjacency matrix, we further threshold $\textbf{A}$ to get the actual edges of the generated graph ({\em i.e.}, every entry above $0.5$ is considered an edge).}

In Table~\ref{tab:aplation_predictor} we show three different configurations of our method: 1)  ``Only Diffusion'': we use the technique we just described, in which the graph is constructed directly from the generated $\mathbf{\Phi}$ and $\boldsymbol{\lambda}$ without using the ``Graph Predictor''; 2) ``Noise + Predictor": we give as input to the ``Graph Predictor" noise drawn from a Gaussian distribution; 3) ``Diff. + Prediction'': this is the full model  used in all the other experiments.  For each configuration, we provide results for two cases: 1) using the full set of eigenpairs - all eigenbasis to train the model, and 2) using a subset of the eigenpairs - model trained  using the 8 largest eigenvalues and their corresponding eigenvectors.


Reconstructing the graph directly from the Laplacian (referred to as ``Only Diffusion'') using the complete spectrum yields the best results. Conversely, using either random noise or the reconstructed Laplacian as input to the Predictor results in significantly inferior outcomes. This suggests that the ``Spectral Diffusion'' part of the network is responsible for the actual generation process. On the other hand, if we consider a truncated eigenbase for the training phase,  the Predictor becomes useful as it helps to refine the results further.

\begin{figure}[h]
    \begin{overpic}[trim=-8mm -5mm 0 -6mm, clip, width=0.65\textwidth]{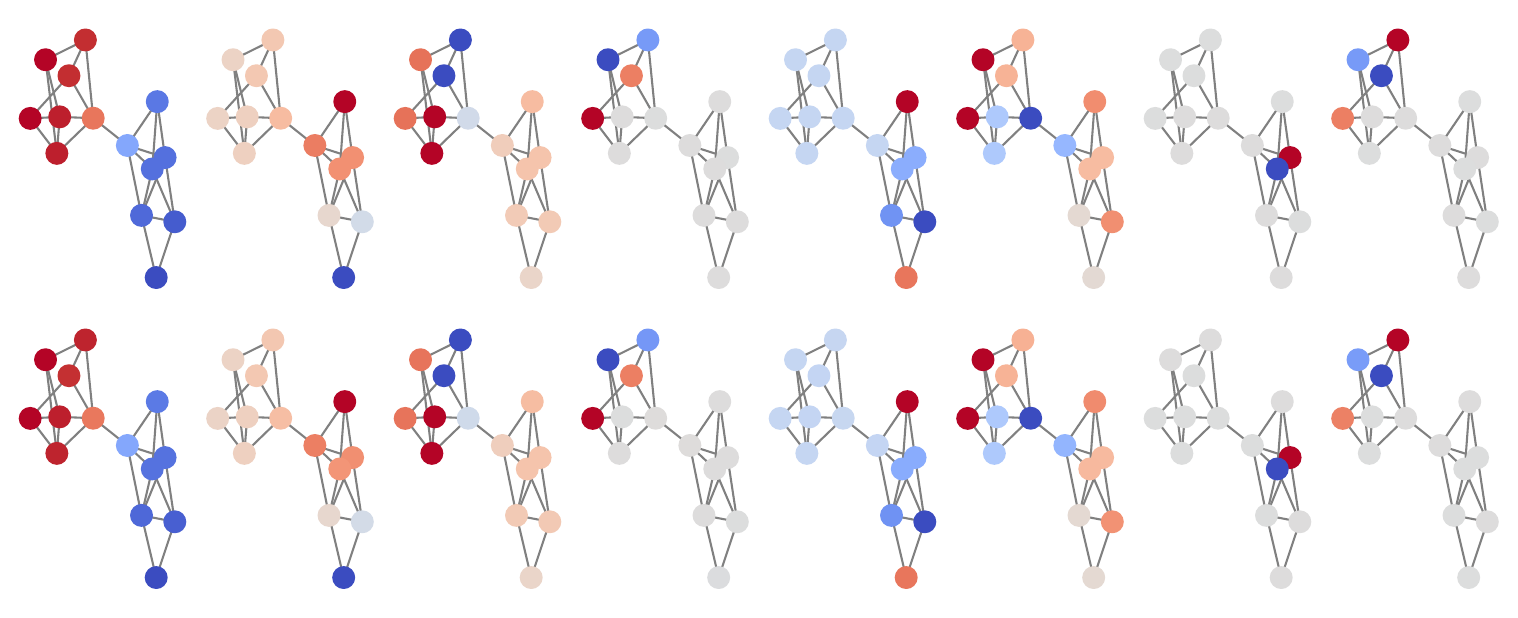}
        \put(0,1){\rotatebox{90}{\scriptsize{Generated $\mathbf{\Phi}_0$}}}
        \put(0,20){\rotatebox{90}{\scriptsize{Computed on $\mathbf{A}$}}}
        \put(8,0){$\phi_1$}
        \put(20.5,0){$\phi_2$}
        \put(33,0){$\phi_3$}
        \put(45.5,0){$\phi_4$}
        \put(58,0){$\phi_5$}
        \put(70.5,0){$\phi_6$}
        \put(83,0){$\phi_7$}
        \put(94.5,0){$\phi_8$}
        
        \put(45,42){\grasp}
        
    \end{overpic}\hfill
    \begin{overpic}[trim=-4mm -3mm 0 -5mm, clip, width=0.33\textwidth]{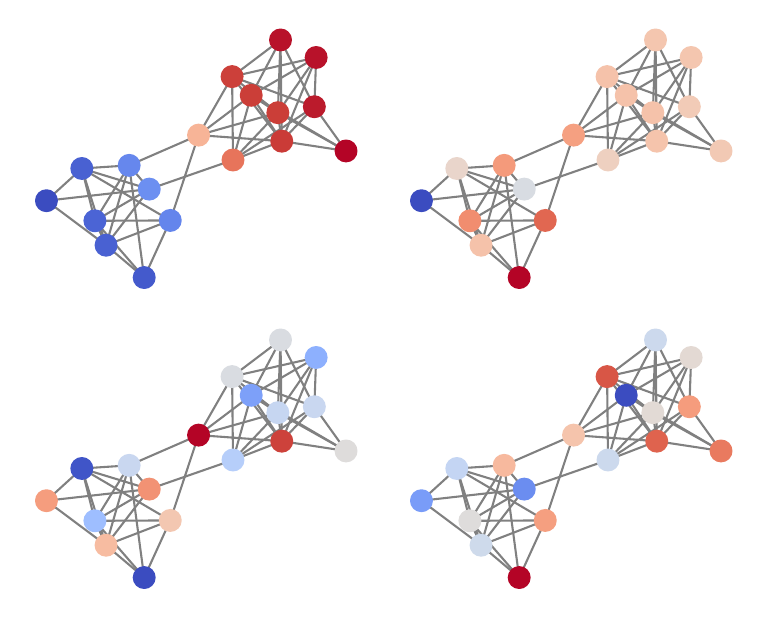}
        \put(0,5){\rotatebox{90}{\scriptsize{Generated $\mathbf{\Phi}$}}}
        \put(0,40){\rotatebox{90}{\scriptsize{Computed on $\mathbf{A}$}}}
        \put(30,0){$\phi_1$}
        \put(70,0){$\phi_2$}
        \put(40,80){SPECTRE}
    \end{overpic}

    \begin{overpic}[trim=-8mm -3mm 0 -5mm, clip, width=0.65\textwidth]{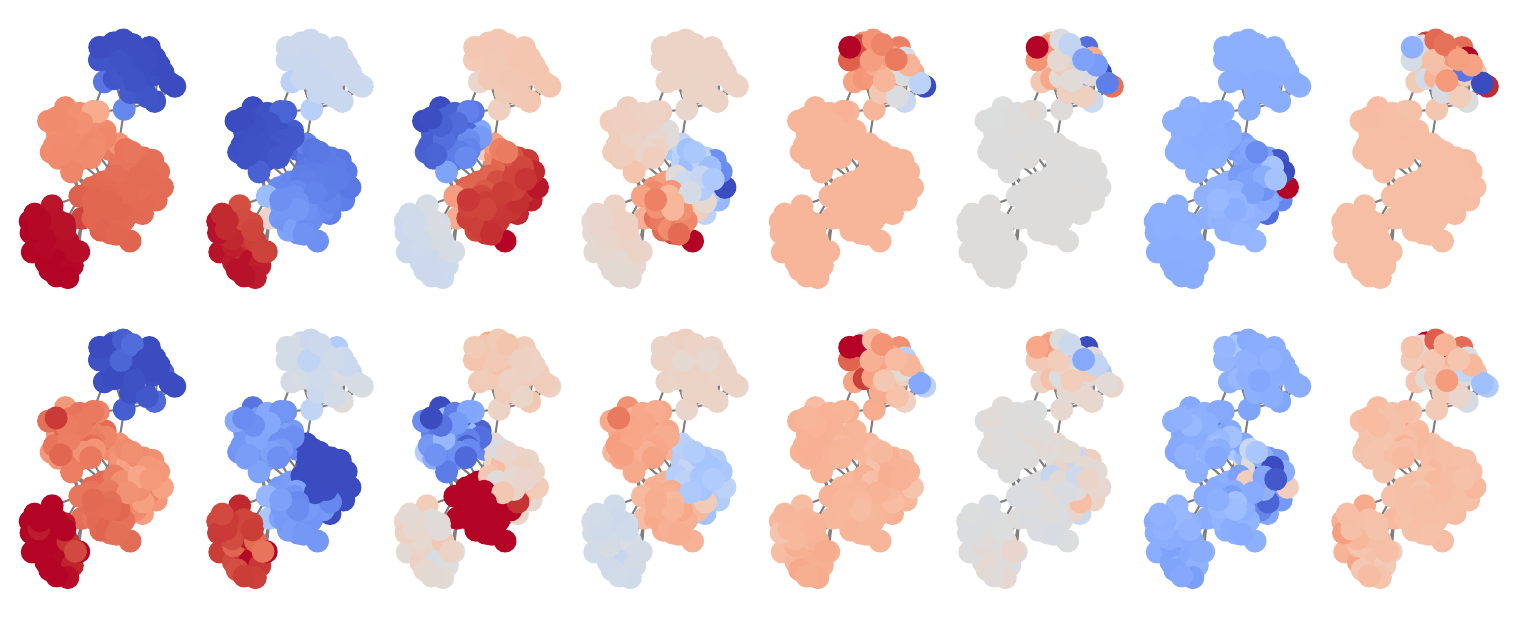}
        \put(0,1){\rotatebox{90}{\scriptsize{Generated $\mathbf{\Phi}_0$}}}
        \put(0,20){\rotatebox{90}{\scriptsize{Computed on $\mathbf{A}$}}}
        \put(5,0){$\phi_1$}
        \put(15.5,0){$\phi_2$}
        \put(28,0){$\phi_3$}
        \put(41,0){$\phi_4$}
        \put(54,0){$\phi_5$}
        \put(66,0){$\phi_6$}
        \put(79,0){$\phi_7$}
        \put(91.5,0){$\phi_8$}        
    \end{overpic}\hfill
    \begin{overpic}[trim=-6mm -3mm 0 -5mm, clip, width=0.33\textwidth]{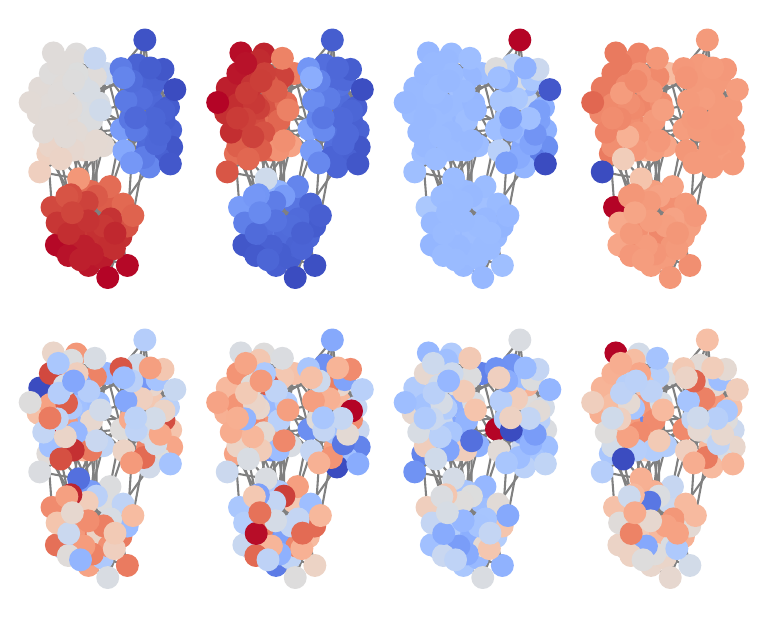}
        \put(0,5){\rotatebox{90}{\scriptsize{Generated $\mathbf{\Phi}$}}}
        \put(0,40){\rotatebox{90}{\scriptsize{Computed on $\mathbf{A}$}}}
        \put(15,0){$\phi_1$}
        \put(35,0){$\phi_2$}
        \put(60,0){$\phi_3$}
        \put(87,0){$\phi_4$}
    \end{overpic}
    \vspace{3mm}
    \caption{\textbf{Left column}: Comparison of the eigenvectors generated by the diffusion module of \grasp~with the eigenvectors recomputed on the Laplacian computed on the adjacency matrix predicted by the PPGN module on the Community (top) and SBM (bottom) datasets. The models have been trained with the 8 smallest eigenvectors. In the smaller dataset (Community) the diffusion generates nearly perfect eigenvectors. In the more challenging SBM dataset we can notice that the generated eigenfunction are slightly different from the one computed on the predicted graph while preserving the overall structure. \textbf{Right column}: comparison of the interpolated eigenvectors that SPECTRE uses to condition the PPGN module to the actual eigenvectors of the generated graph. In this case the eigenvectors structure is completely lost in the generative process.}
    \label{fig:evecs_consistency}
\end{figure}

\red{To provide a comprehensive overview,  we  also provide some qualitative examples in \Cref{fig:evecs_consistency}, comparing the eigenvectors generated by the diffusion module of \grasp~with those of the Laplacian computed on the adjacency matrix predicted by the PPGN module. In smaller graphs, the eigenvectors are almost perfectly preserved, while only minor local differences emerge in larger graphs. In contrast, the generative approach used by SPECTRE fails to maintain the relationship between the conditioning eigenvectors and the final generated graph.}

\begin{table}[b!]\centering
\caption{Ablation of the Graph Predictor network on the community-small dataset\label{tab:aplation_predictor}.}
\scriptsize
\begin{tabular}{lcccc|cccc}\toprule
&\multicolumn{4}{c}{Full Set of Eigenpairs} &\multicolumn{4}{c}{First 8 Eigenpairs}\\
\cmidrule{2-9}
&Deg. $\downarrow$ &Clus. $\downarrow$ &Spect. $\downarrow$ &Orb. $\downarrow$ &Deg. $\downarrow$ &Clus. $\downarrow$ &Spect. $\downarrow$ &Orb. $\downarrow$ \\
\midrule 
\midrule 
Only Diffusion &\textbf{0.00134} &\textbf{0.05484} &\textbf{ 0.01574} &\textbf{0.00514 } &0.00936 &0.09753 &0.04341 &0.01255 \\
Noise + Predictor &0.04428 &0.11129 &0.05691 &0.46051 &0.06681 &0.12330 &0.05972 &0.45137 \\
Diff. + Predictor &0.00562 &0.07887 &0.02232 &0.00942 &\textbf{0.00423} &\textbf{0.05464} &\textbf{0.01832} &\textbf{0.00866} \\
\bottomrule
\end{tabular}
\end{table}

\subsection{Eigenvectors Orthogonality}

\red{The general framework of DDPM cannot guarantee the orthonormality of the generated eigenvectors. While in principle this might pose a problem, in practice we observed that this property is well preserved in the generations. In \Cref{sec:apportho}, we present both quantitative and qualitative analyses to evaluate the orthogonality of the generated eigenvectors, demonstrating that they exhibit approximate orthonormality. Moreover, in order to test if having an exact orthonormality of the eigenvectors brings any advantage, we tried to reproject the final generated eigenvectors to an orthonormal matrix through QR decomposition before the PPGN predictor step. As reported in \Cref{fig:comparison_ortho}, we did not observe a clear benefit in having orthonormal basis. We argue that, even if orthonormality is a well-known characteristic of the eigendecomposition of the graph Laplacian (indeed, of the eigendecomposition of \revised{any symmetric} matrix), it is probably not the most important (nor essential) to guarantee a good reconstruction of a valid graph Laplacian, which exhibits more complex properties that need to be learned directly from data.}


\subsection{Spectral Conditioned Generation}\label{subsec:conditioning}
The spectrum of the Laplacian plays an important role in many applications, from graph classification and mesh analysis~\cite{cosmo2020amks, cosmo20223d,bai2015quantum,hu2014stable, minello2019neumann,gasparetto2015non,gasparetto2015nonshape} to reconstructing the underlying geometry of a triangulated 3D shape~\cite{isospec,marin2021,moschella2022learning} and to define universal adversarial attacks~\cite{rampini2021universal}. Being able to generate a graph given a spectrum is thus an important feature of a generative method. We pose the graph generation problem conditioned on a sequence of eigenvalues and/or eigenvectors as an inpainting problem~\cite{lugmayr2022repaint}.
\begin{figure*}[t!]
\centering
\includegraphics[width=0.42\textwidth]{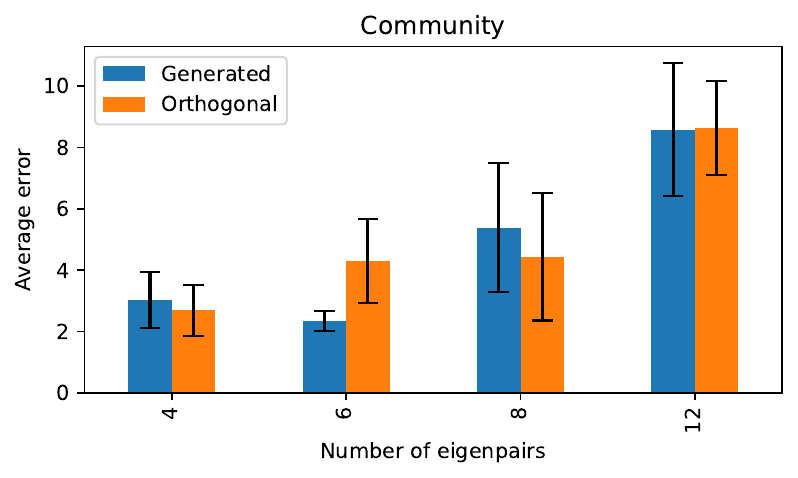}
\includegraphics[width=0.415\textwidth]{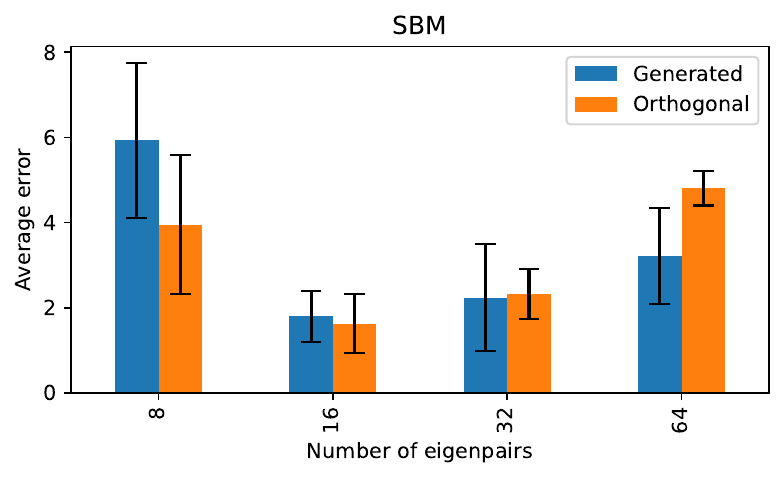}
\caption{
Performance analysis without (Generated) and with (Orthonormal) reprojecting the generated eigenvectors to an orthonormal basis. The average error represents the mean degradation of metrics between the generated graphs and the training set. We report both the mean and the standard deviation as error bars on 10 generations of 200 graphs. Specifically, Degree, Cluster, and Spectral metrics are calculated between the generated graphs and the test set, then normalized by the metrics between the training and test sets.}
\label{fig:comparison_ortho}  
\end{figure*}

\paragraph{Eigenvalues conditioned generation} In this setup, the eigenvectors at time $t-1$ are computed according to Eq.~\ref{eq:denoise_step}, while the eigenvalues are derived from the target ones through the diffusion process of Eq.~\ref{eq:forward_diffusion_xt_x0}, {\em i.e.},
\begin{equation}
\begin{aligned}
\boldsymbol{\lambda}_{t-1} & \sim \mathcal{N}\left(\sqrt{\bar{\alpha}_t} \boldsymbol{\lambda}_0,\left(1-\bar{\alpha}_t\right) \mathbf{I}\right), \qquad
\mathbf{\Phi}_{t-1} & \sim \mathcal{N}\left(\mathbf{\Phi}_{t-1}; \boldsymbol{\mu}_{\theta}\left(\mathbf{\Phi}_t, \boldsymbol{\lambda}_t; t \right), \beta_t \mathbf{I} \right)\;. \\
\end{aligned}
\end{equation}

To validate the generation conditioned on the eigenvalues, we generate graphs of the SBM data distribution by fixing the number of nodes as the average number of nodes of graphs containing 2 and 3 communities (70 nodes). \revised{We randomly choose one graph with 2 communities and one graph with 3 communities from the test set, and we consider their spectra. We use these to condition the generation of two sets of 100 graphs, for the 2 and 3 communities eigenvalue sequences, respectively.} The results reported in  Table~\ref{tab:gen_comm_table} show that spectral conditioning is able to influence the properties of the generated graphs, while the spectral conditioning provided by SPECTRE fails to do so.

\paragraph{Eigenvectors conditioned generation} We adopt a similar strategy to condition on a subset of the eigenvectors. In this case, the portion of $k^\prime$ known eigenvectors $\mathbf{\Phi}^\prime = [\mathbf{\phi}_0,\dots \mathbf{\phi}_{k^\prime}]$ at time $t-1$ is computed according to Eq.~\ref{eq:denoise_step}, while the remaining eigenvectors $\overline{\mathbf{\Phi}^\prime} = [\mathbf{\phi}_{k^\prime+1},\dots \mathbf{\phi}_{k}]$ and the eigenvalues are derived from the target ones through the forward diffusion process. The three groups of 4 graphs in Figure~\ref{fig:evec_conditioning} show the donor graphs (left) from which the first three eigenvectors were computed, the graphs conditioned on the given eigenvectors using \grasp~(center) and SPECTRE (right). The color is the 2D color encoding of 2 of the three first eigenvectors manually selected to highlight the different clusters. The colors from the donor graph have then been transported on the generated graphs' corresponding node (same node index). While \grasp~is able to preserve the community structure encoded by the given eigenvectors, in SPECTRE this information is completely lost, causing nodes from the same community in the donor graph to randomly spread over different communities of the generated graph. \revised{This may be due to the particular generation mechanism of SPECTRE. Specifically, SPECTRE learns a set of reduced orthogonal bases during training, which are left and right-rotated according to a rotation matrix predicted by a PointNetST \cite{segoluniversal} network based on some input (generated) eigenvalues. This requires an alignment of the graphs to the learned bases, which makes the training more complex. Our approach on the other hand is fully covariant.}

\begin{figure*}[t!]
\centering 
\captionlistentry[figure]{Eigenvectors conditioning} \label{fig:evec_conditioning}  
\begin{minipage}[t]{.75\linewidth}
    \begin{overpic}[clip, width=0.99\textwidth]{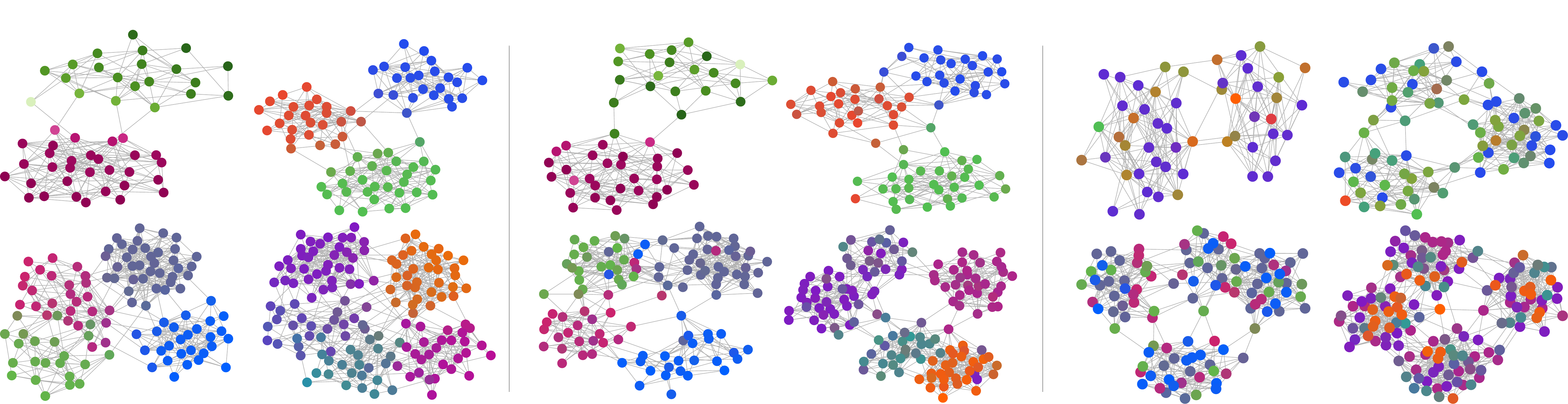}
        \put(16.5,25){\makebox(0,0){\scriptsize Donor Graph}}
        \put(50,25){\makebox(0,0){\scriptsize \grasp}}
        \put(82.5,25){\makebox(0,0){\scriptsize SPECTRE}}
    \end{overpic}
\end{minipage}%
\hfill
\scriptsize
\def\arraystretch{0.8}
\captionlistentry[table]{Eigenvalues conditioning}
\label{tab:gen_comm_table}
\begin{minipage}[t]{.22\linewidth}
\scriptsize
\setlength{\tabcolsep}{3pt}
\begin{tabular}[b]{lccc} 
\toprule  
\# gen. comm.: &  2 &  3 & 4   \\
\midrule \midrule
\grasp\\
\midrule 
$\lambda$ of 2 comm. & \textbf{76} & 19 & 5  \\
$\lambda$ of 3 comm. & 12 & \textbf{88} & 0  \\
\midrule \midrule
SPECTRE\\
\midrule 
$\lambda$ of 2 comm. & \textbf{93} & 7 & 0  \\
$\lambda$ of 3 comm. & 96 & \textbf{4} & 0  \\
\bottomrule 
\end{tabular} 
\end{minipage} 
\captionlistentry[table]{This is a table.}
 \addtocounter{table}{-1}
  \addtocounter{figure}{-1}
\captionsetup{labelformat=andtable}
\caption{
Conditioning of the generation on the first 3 smallest eigenvectors as an inpainting task using RePaint \cite{lugmayr2022repaint} (Figure). Number of communities in the graphs generated with spectrum conditioning (Table). Higher values should appear in the bold diagonal.}

\end{figure*}


\section{Conclusion}\label{sec:conclusion}
We have introduced \grasp, a diffusion-based generative model for graphs where the spectrum of the graph Laplacian is used to retain structural information while reducing the computational complexity. Our approach has a number of advantages, from the ability to directly generate eigenvectors and eigenvalues, to the possibility of naturally encapsulate node feature information as well as conditioning the generation on target spectral properties.

Our model suffers from two main limitations. Firstly, while using low/high frequencies to reconstruct the spectrum is theoretically grounded, in practice it would be interesting to allow the model to select the most informative frequencies for a given dataset. We will explore this possibility in future work. Secondly, while the diffusion model is linear, the bottleneck of our model is the PPGN-based predictor, which has quadratic complexity. In the future, it would be interesting to investigate alternative methods using a sparse representation of the adjacency matrix reconstructed from the Laplacian.

\section*{Acknowledgments}
 L.C. and A.B. are supported by the PRIN 2022 project n. 2022AL45R2 (EYE-FI.AI, CUP H53D2300350-0001). G.M. acknowledges financial support from the European Union - NextGenerationEU, in the framework of the iNEST - Interconnected Nord-Est Innovation Ecosystem (iNEST ECS$\_$00000043 – CUP H43C22000540006). In this regard, the views and opinions expressed are solely those of the authors and do not necessarily reflect those of the European Union, nor can the European Union be held responsible for them. 


\bibliography{biblio}

\begin{thebibliography}{64}
\providecommand{\natexlab}[1]{#1}
\providecommand{\url}[1]{\texttt{#1}}
\expandafter\ifx\csname urlstyle\endcsname\relax
  \providecommand{\doi}[1]{doi: #1}\else
  \providecommand{\doi}{doi: \begingroup \urlstyle{rm}\Url}\fi

\bibitem[Albert \& Barab{\'a}si(2002)Albert and Barab{\'a}si]{albert2002statistical}
R{\'e}ka Albert and Albert-L{\'a}szl{\'o} Barab{\'a}si.
\newblock Statistical mechanics of complex networks.
\newblock \emph{Reviews of modern physics}, 74\penalty0 (1):\penalty0 47, 2002.

\bibitem[Bai et~al.(2015)Bai, Rossi, Torsello, and Hancock]{bai2015quantum}
Lu~Bai, Luca Rossi, Andrea Torsello, and Edwin~R Hancock.
\newblock A quantum jensen--shannon graph kernel for unattributed graphs.
\newblock \emph{Pattern Recognition}, 48\penalty0 (2):\penalty0 344--355, 2015.

\bibitem[Chung(1997)]{chung1997spectral}
Fan~RK Chung.
\newblock \emph{Spectral graph theory}, volume~92.
\newblock American Mathematical Soc., 1997.

\bibitem[Cosmo et~al.(2016)Cosmo, Rodola, Masci, Torsello, and Bronstein]{cosmo2016matching}
Luca Cosmo, Emanuele Rodola, Jonathan Masci, Andrea Torsello, and Michael~M Bronstein.
\newblock Matching deformable objects in clutter.
\newblock In \emph{2016 Fourth international conference on 3D vision (3DV)}, pp.\  1--10. IEEE, 2016.

\bibitem[Cosmo et~al.(2019)Cosmo, Panine, Rampini, Ovsjanikov, Bronstein, and Rodol{\`{a}}]{isospec}
Luca Cosmo, Mikhail Panine, Arianna Rampini, Maks Ovsjanikov, Michael~M. Bronstein, and Emanuele Rodol{\`{a}}.
\newblock Isospectralization, or how to hear shape, style, and correspondence.
\newblock In \emph{{IEEE} Conference on Computer Vision and Pattern Recognition, {CVPR} 2019, Long Beach, CA, USA, June 16-20, 2019}, pp.\  7529--7538. Computer Vision Foundation / {IEEE}, 2019.
\newblock \doi{10.1109/CVPR.2019.00771}.

\bibitem[Cosmo et~al.(2020)Cosmo, Minello, Bronstein, Rossi, and Torsello]{cosmo2020amks}
Luca Cosmo, Giorgia Minello, Michael Bronstein, Luca Rossi, and Andrea Torsello.
\newblock The average mixing kernel signature.
\newblock In Andrea Vedaldi, Horst Bischof, Thomas Brox, and Jan-Michael Frahm (eds.), \emph{Computer Vision -- ECCV 2020}, pp.\  1--17, Cham, 2020. Springer International Publishing.
\newblock ISBN 978-3-030-58565-5.

\bibitem[Cosmo et~al.(2022)Cosmo, Minello, Bronstein, Rodol{\`a}, Rossi, and Torsello]{cosmo20223d}
Luca Cosmo, Giorgia Minello, Michael Bronstein, Emanuele Rodol{\`a}, Luca Rossi, and Andrea Torsello.
\newblock 3d shape analysis through a quantum lens: the average mixing kernel signature.
\newblock \emph{International Journal of Computer Vision}, 130\penalty0 (6):\penalty0 1474--1493, 2022.

\bibitem[De~Cao \& Kipf(2018)De~Cao and Kipf]{de2018molgan}
Nicola De~Cao and Thomas Kipf.
\newblock Molgan: An implicit generative model for small molecular graphs.
\newblock \emph{arXiv preprint arXiv:1805.11973}, 2018.

\bibitem[{de Fraysseix} \& {Ossona de Mendez}(2012){de Fraysseix} and {Ossona de Mendez}]{DEFRAYSSEIX2012279}
Hubert {de Fraysseix} and Patrice {Ossona de Mendez}.
\newblock Trémaux trees and planarity.
\newblock \emph{European Journal of Combinatorics}, 33\penalty0 (3):\penalty0 279--293, 2012.
\newblock ISSN 0195-6698.
\newblock \doi{https://doi.org/10.1016/j.ejc.2011.09.012}.
\newblock URL \url{https://www.sciencedirect.com/science/article/pii/S0195669811001600}.
\newblock Topological and Geometric Graph Theory.

\bibitem[Dobson \& Doig(2003)Dobson and Doig]{dobson2003distinguishing}
Paul~D Dobson and Andrew~J Doig.
\newblock Distinguishing enzyme structures from non-enzymes without alignments.
\newblock \emph{Journal of molecular biology}, 330\penalty0 (4):\penalty0 771--783, 2003.

\bibitem[Drobyshevskiy \& Turdakov(2019)Drobyshevskiy and Turdakov]{drobyshevskiy2019random}
Mikhail Drobyshevskiy and Denis Turdakov.
\newblock Random graph modeling: A survey of the concepts.
\newblock \emph{ACM computing surveys (CSUR)}, 52\penalty0 (6):\penalty0 1--36, 2019.

\bibitem[Erd{\H{o}}s et~al.(1960)Erd{\H{o}}s, R{\'e}nyi, et~al.]{erdHos1960evolution}
Paul Erd{\H{o}}s, Alfr{\'e}d R{\'e}nyi, et~al.
\newblock On the evolution of random graphs.
\newblock \emph{Publ. math. inst. hung. acad. sci}, 5\penalty0 (1):\penalty0 17--60, 1960.

\bibitem[Fahrmeir et~al.(2013)Fahrmeir, Kneib, Lang, Marx, Fahrmeir, Kneib, Lang, and Marx]{fahrmeir2013regression}
Ludwig Fahrmeir, Thomas Kneib, Stefan Lang, Brian Marx, Ludwig Fahrmeir, Thomas Kneib, Stefan Lang, and Brian Marx.
\newblock \emph{Regression models}.
\newblock Springer, 2013.

\bibitem[Faloutsos(2008)]{faloutsos2008graph}
Christos Faloutsos.
\newblock Graph mining: Laws, generators and tools.
\newblock \emph{Lecture Notes in Computer Science}, 5012:\penalty0 1, 2008.

\bibitem[Feller(1949)]{feller49}
W.~Feller.
\newblock On the theory of stochastic processes, with particular reference to applications.
\newblock In \emph{Proceedings of the [First] Berkeley Symposium on Mathematical Statistics and Probability}, pp.\  403--432, 1949.

\bibitem[Gasparetto et~al.(2015{\natexlab{a}})Gasparetto, Minello, and Torsello]{gasparetto2015non}
Andrea Gasparetto, Giorgia Minello, and Andrea Torsello.
\newblock A non-parametric spectral model for graph classification.
\newblock In \emph{International Conference on Pattern Recognition Applications and Methods}, volume~2, pp.\  312--319. SciTePress, 2015{\natexlab{a}}.

\bibitem[Gasparetto et~al.(2015{\natexlab{b}})Gasparetto, Minello, and Torsello]{gasparetto2015nonshape}
Andrea Gasparetto, Giorgia Minello, and Andrea Torsello.
\newblock Non-parametric spectral model for shape retrieval.
\newblock In \emph{2015 International Conference on 3D Vision}, pp.\  344--352. IEEE, 2015{\natexlab{b}}.

\bibitem[G{\'o}mez-Bombarelli et~al.(2018)G{\'o}mez-Bombarelli, Wei, Duvenaud, Hern{\'a}ndez-Lobato, S{\'a}nchez-Lengeling, Sheberla, Aguilera-Iparraguirre, Hirzel, Adams, and Aspuru-Guzik]{gomez2018automatic}
Rafael G{\'o}mez-Bombarelli, Jennifer~N Wei, David Duvenaud, Jos{\'e}~Miguel Hern{\'a}ndez-Lobato, Benjam{\'\i}n S{\'a}nchez-Lengeling, Dennis Sheberla, Jorge Aguilera-Iparraguirre, Timothy~D Hirzel, Ryan~P Adams, and Al{\'a}n Aspuru-Guzik.
\newblock Automatic chemical design using a data-driven continuous representation of molecules.
\newblock \emph{ACS central science}, 4\penalty0 (2):\penalty0 268--276, 2018.

\bibitem[Goodfellow et~al.(2014)Goodfellow, Pouget-Abadie, Mirza, Xu, Warde-Farley, Ozair, Courville, and Bengio]{goodfellow2014generative}
Ian Goodfellow, Jean Pouget-Abadie, Mehdi Mirza, Bing Xu, David Warde-Farley, Sherjil Ozair, Aaron Courville, and Yoshua Bengio.
\newblock Generative adversarial nets.
\newblock \emph{Advances in neural information processing systems}, 27, 2014.

\bibitem[Grover et~al.(2019)Grover, Zweig, and Ermon]{grover2019graphite}
Aditya Grover, Aaron Zweig, and Stefano Ermon.
\newblock Graphite: Iterative generative modeling of graphs.
\newblock In \emph{International conference on machine learning}, pp.\  2434--2444. PMLR, 2019.

\bibitem[Guo \& Zhao(2022)Guo and Zhao]{guo2022systematic}
Xiaojie Guo and Liang Zhao.
\newblock A systematic survey on deep generative models for graph generation.
\newblock \emph{IEEE Transactions on Pattern Analysis and Machine Intelligence}, 45\penalty0 (5):\penalty0 5370--5390, 2022.

\bibitem[Haefeli et~al.(2022)Haefeli, Martinkus, Perraudin, and Wattenhofer]{haefeli2022diffusion}
Kilian~Konstantin Haefeli, Karolis Martinkus, Nathana{\"e}l Perraudin, and Roger Wattenhofer.
\newblock Diffusion models for graphs benefit from discrete state spaces.
\newblock \emph{arXiv preprint arXiv:2210.01549}, 2022.

\bibitem[Ho et~al.(2020)Ho, Jain, and Abbeel]{ho2020denoising}
Jonathan Ho, Ajay Jain, and Pieter Abbeel.
\newblock Denoising diffusion probabilistic models.
\newblock \emph{Advances in neural information processing systems}, 33:\penalty0 6840--6851, 2020.

\bibitem[Hu et~al.(2014)Hu, Rustamov, and Guibas]{hu2014stable}
Nan Hu, Raif~M Rustamov, and Leonidas Guibas.
\newblock Stable and informative spectral signatures for graph matching.
\newblock In \emph{Proceedings of the IEEE conference on computer vision and pattern recognition}, pp.\  2305--2312, 2014.

\bibitem[Huang et~al.(2022)Huang, Sun, Du, Fu, and Lv]{huang2022graphgdp}
Han Huang, Leilei Sun, Bowen Du, Yanjie Fu, and Weifeng Lv.
\newblock Graphgdp: Generative diffusion processes for permutation invariant graph generation.
\newblock In \emph{2022 IEEE International Conference on Data Mining (ICDM)}, pp.\  201--210. IEEE, 2022.

\bibitem[Jo et~al.(2022)Jo, Lee, and Hwang]{jo2022score}
Jaehyeong Jo, Seul Lee, and Sung~Ju Hwang.
\newblock Score-based generative modeling of graphs via the system of stochastic differential equations.
\newblock In \emph{International Conference on Machine Learning}, pp.\  10362--10383. PMLR, 2022.

\bibitem[Kingma \& Welling(2013)Kingma and Welling]{kingma2013auto}
Diederik~P Kingma and Max Welling.
\newblock Auto-encoding variational bayes.
\newblock \emph{arXiv preprint arXiv:1312.6114}, 2013.

\bibitem[Krawczuk et~al.(2020)Krawczuk, Abranches, Loukas, and Cevher]{krawczuk2020gg}
Igor Krawczuk, Pedro Abranches, Andreas Loukas, and Volkan Cevher.
\newblock Gg-gan: A geometric graph generative adversarial network.
\newblock 2020.

\bibitem[Leskovec \& Faloutsos(2007)Leskovec and Faloutsos]{leskovec2007scalable}
Jure Leskovec and Christos Faloutsos.
\newblock Scalable modeling of real graphs using kronecker multiplication.
\newblock In \emph{Proceedings of the 24th international conference on Machine learning}, pp.\  497--504, 2007.

\bibitem[Leskovec et~al.(2010)Leskovec, Chakrabarti, Kleinberg, Faloutsos, and Ghahramani]{leskovec2010kronecker}
Jure Leskovec, Deepayan Chakrabarti, Jon Kleinberg, Christos Faloutsos, and Zoubin Ghahramani.
\newblock Kronecker graphs: an approach to modeling networks.
\newblock \emph{Journal of Machine Learning Research}, 11\penalty0 (2), 2010.

\bibitem[Li et~al.(2018{\natexlab{a}})Li, Zhang, and Liu]{li2018multi}
Yibo Li, Liangren Zhang, and Zhenming Liu.
\newblock Multi-objective de novo drug design with conditional graph generative model.
\newblock \emph{Journal of cheminformatics}, 10:\penalty0 1--24, 2018{\natexlab{a}}.

\bibitem[Li et~al.(2018{\natexlab{b}})Li, Vinyals, Dyer, Pascanu, and Battaglia]{li2018learning}
Yujia Li, Oriol Vinyals, Chris Dyer, Razvan Pascanu, and Peter Battaglia.
\newblock Learning deep generative models of graphs.
\newblock \emph{arXiv preprint arXiv:1803.03324}, 2018{\natexlab{b}}.

\bibitem[Liao et~al.(2019)Liao, Li, Song, Wang, Hamilton, Duvenaud, Urtasun, and Zemel]{liao2019efficient}
Renjie Liao, Yujia Li, Yang Song, Shenlong Wang, Will Hamilton, David~K Duvenaud, Raquel Urtasun, and Richard Zemel.
\newblock Efficient graph generation with graph recurrent attention networks.
\newblock \emph{Advances in neural information processing systems}, 32, 2019.

\bibitem[Lugmayr et~al.(2022)Lugmayr, Danelljan, Romero, Yu, Timofte, and Van~Gool]{lugmayr2022repaint}
Andreas Lugmayr, Martin Danelljan, Andres Romero, Fisher Yu, Radu Timofte, and Luc Van~Gool.
\newblock Repaint: Inpainting using denoising diffusion probabilistic models.
\newblock 2022.

\bibitem[Luo et~al.(2003)Luo, Wilson, and Hancock]{luo2003spectral}
Bin Luo, Richard~C Wilson, and Edwin~R Hancock.
\newblock Spectral embedding of graphs.
\newblock \emph{Pattern recognition}, 36\penalty0 (10):\penalty0 2213--2230, 2003.

\bibitem[Luo et~al.(2023)Luo, Mo, and Pan]{luo2023fast}
Tianze Luo, Zhanfeng Mo, and Sinno~Jialin Pan.
\newblock Fast graph generation via spectral diffusion.
\newblock \emph{IEEE Transactions on Pattern Analysis and Machine Intelligence}, 2023.

\bibitem[Luo et~al.(2021)Luo, Yan, and Ji]{luo2021graphdf}
Youzhi Luo, Keqiang Yan, and Shuiwang Ji.
\newblock Graphdf: A discrete flow model for molecular graph generation.
\newblock In \emph{International Conference on Machine Learning}, pp.\  7192--7203. PMLR, 2021.

\bibitem[Marin et~al.(2021)Marin, Rampini, Castellani, Rodol\`{a}, Ovsjanikov, and Melzi]{marin2021}
Riccardo Marin, Arianna Rampini, Umberto Castellani, Emanuele Rodol\`{a}, Maks Ovsjanikov, and Simone Melzi.
\newblock Spectral shape recovery and analysis via data-driven connections.
\newblock \emph{Int. J. Comput. Vision}, 129\penalty0 (10):\penalty0 2745–2760, oct 2021.
\newblock ISSN 0920-5691.
\newblock \doi{10.1007/s11263-021-01492-6}.

\bibitem[Maron et~al.(2019)Maron, Ben-Hamu, Serviansky, and Lipman]{NEURIPS2019_bb04af0f}
Haggai Maron, Heli Ben-Hamu, Hadar Serviansky, and Yaron Lipman.
\newblock Provably powerful graph networks.
\newblock In H.~Wallach, H.~Larochelle, A.~Beygelzimer, F.~d\textquotesingle Alch\'{e}-Buc, E.~Fox, and R.~Garnett (eds.), \emph{Advances in Neural Information Processing Systems}, volume~32. Curran Associates, Inc., 2019.
\newblock URL \url{https://proceedings.neurips.cc/paper_files/paper/2019/file/bb04af0f7ecaee4aae62035497da1387-Paper.pdf}.

\bibitem[Martinkus et~al.(2022)Martinkus, Loukas, Perraudin, and Wattenhofer]{martinkus2022spectre}
Karolis Martinkus, Andreas Loukas, Nathana{\"e}l Perraudin, and Roger Wattenhofer.
\newblock Spectre: Spectral conditioning helps to overcome the expressivity limits of one-shot graph generators.
\newblock In \emph{International Conference on Machine Learning}, pp.\  15159--15179. PMLR, 2022.

\bibitem[Minello et~al.(2019)Minello, Rossi, and Torsello]{minello2019neumann}
Giorgia Minello, Luca Rossi, and Andrea Torsello.
\newblock On the von neumann entropy of graphs.
\newblock \emph{Journal of Complex Networks}, 7\penalty0 (4):\penalty0 491--514, 2019.

\bibitem[Moschella et~al.(2022)Moschella, Melzi, Cosmo, Maggioli, Litany, Ovsjanikov, Guibas, and Rodol{\`a}]{moschella2022learning}
Luca Moschella, Simone Melzi, Luca Cosmo, Filippo Maggioli, Or~Litany, Maks Ovsjanikov, Leonidas Guibas, and Emanuele Rodol{\`a}.
\newblock Learning spectral unions of partial deformable 3d shapes.
\newblock In \emph{Computer Graphics Forum}, volume~41, pp.\  407--417. Wiley Online Library, 2022.

\bibitem[Niu et~al.(2020)Niu, Song, Song, Zhao, Grover, and Ermon]{niu2020permutation}
Chenhao Niu, Yang Song, Jiaming Song, Shengjia Zhao, Aditya Grover, and Stefano Ermon.
\newblock Permutation invariant graph generation via score-based generative modeling.
\newblock In \emph{International Conference on Artificial Intelligence and Statistics}, pp.\  4474--4484. PMLR, 2020.

\bibitem[Peixoto(2014)]{sbmgreedy}
Tiago~P. Peixoto.
\newblock Hierarchical block structures and high-resolution model selection in large networks.
\newblock \emph{Phys. Rev. X}, 4:\penalty0 011047, Mar 2014.
\newblock \doi{10.1103/PhysRevX.4.011047}.
\newblock URL \url{https://link.aps.org/doi/10.1103/PhysRevX.4.011047}.

\bibitem[Popova et~al.(2019)Popova, Shvets, Oliva, and Isayev]{popova2019molecularrnn}
Mariya Popova, Mykhailo Shvets, Junier Oliva, and Olexandr Isayev.
\newblock Molecularrnn: Generating realistic molecular graphs with optimized properties.
\newblock \emph{arXiv preprint arXiv:1905.13372}, 2019.

\bibitem[Ramakrishnan et~al.(2014)Ramakrishnan, Dral, Rupp, and Von~Lilienfeld]{ramakrishnan2014quantum}
Raghunathan Ramakrishnan, Pavlo~O Dral, Matthias Rupp, and O~Anatole Von~Lilienfeld.
\newblock Quantum chemistry structures and properties of 134 kilo molecules.
\newblock \emph{Scientific data}, 1\penalty0 (1):\penalty0 1--7, 2014.

\bibitem[Rampini et~al.(2021)Rampini, Pestarini, Cosmo, Melzi, and Rodola]{rampini2021universal}
Arianna Rampini, Franco Pestarini, Luca Cosmo, Simone Melzi, and Emanuele Rodola.
\newblock Universal spectral adversarial attacks for deformable shapes.
\newblock In \emph{Proceedings of the IEEE/CVF conference on computer vision and pattern recognition}, pp.\  3216--3226, 2021.

\bibitem[Rodol{\`a} et~al.(2017)Rodol{\`a}, Cosmo, Bronstein, Torsello, and Cremers]{rodola2017partial}
Emanuele Rodol{\`a}, Luca Cosmo, Michael~M Bronstein, Andrea Torsello, and Daniel Cremers.
\newblock Partial functional correspondence.
\newblock In \emph{Computer graphics forum}, volume~36, pp.\  222--236. Wiley Online Library, 2017.

\bibitem[Ruddigkeit et~al.(2012)Ruddigkeit, Van~Deursen, Blum, and Reymond]{ruddigkeit2012enumeration}
Lars Ruddigkeit, Ruud Van~Deursen, Lorenz~C Blum, and Jean-Louis Reymond.
\newblock Enumeration of 166 billion organic small molecules in the chemical universe database gdb-17.
\newblock \emph{Journal of chemical information and modeling}, 52\penalty0 (11):\penalty0 2864--2875, 2012.

\bibitem[Samanta et~al.(2020)Samanta, De, Jana, G{\'o}mez, Chattaraj, Ganguly, and Gomez-Rodriguez]{samanta2020nevae}
Bidisha Samanta, Abir De, Gourhari Jana, Vicen{\c{c}} G{\'o}mez, Pratim Chattaraj, Niloy Ganguly, and Manuel Gomez-Rodriguez.
\newblock Nevae: A deep generative model for molecular graphs.
\newblock \emph{Journal of machine learning research}, 21\penalty0 (114):\penalty0 1--33, 2020.

\bibitem[Segol \& Lipman(2019)Segol and Lipman]{segoluniversal}
Nimrod Segol and Yaron Lipman.
\newblock On universal equivariant set networks.
\newblock In \emph{International Conference on Learning Representations}, 2019.

\bibitem[Shi et~al.(2020)Shi, Xu, Zhu, Zhang, Zhang, and Tang]{shi2020graphaf}
Chence Shi, Minkai Xu, Zhaocheng Zhu, Weinan Zhang, Ming Zhang, and Jian Tang.
\newblock Graphaf: a flow-based autoregressive model for molecular graph generation.
\newblock \emph{arXiv preprint arXiv:2001.09382}, 2020.

\bibitem[Simonovsky \& Komodakis(2018)Simonovsky and Komodakis]{simonovsky2018graphvae}
Martin Simonovsky and Nikos Komodakis.
\newblock Graphvae: Towards generation of small graphs using variational autoencoders.
\newblock In \emph{Artificial Neural Networks and Machine Learning--ICANN 2018: 27th International Conference on Artificial Neural Networks, Rhodes, Greece, October 4-7, 2018, Proceedings, Part I 27}, pp.\  412--422. Springer, 2018.

\bibitem[Sohl-Dickstein et~al.(2015)Sohl-Dickstein, Weiss, Maheswaranathan, and Ganguli]{sohl2015deep}
Jascha Sohl-Dickstein, Eric Weiss, Niru Maheswaranathan, and Surya Ganguli.
\newblock Deep unsupervised learning using nonequilibrium thermodynamics.
\newblock In \emph{International conference on machine learning}, pp.\  2256--2265. PMLR, 2015.

\bibitem[Song \& Ermon(2019)Song and Ermon]{song2019generative}
Yang Song and Stefano Ermon.
\newblock Generative modeling by estimating gradients of the data distribution.
\newblock \emph{Advances in neural information processing systems}, 32, 2019.

\bibitem[Vaswani et~al.(2017)Vaswani, Shazeer, Parmar, Uszkoreit, Jones, Gomez, Kaiser, and Polosukhin]{NIPS2017_3f5ee243}
Ashish Vaswani, Noam Shazeer, Niki Parmar, Jakob Uszkoreit, Llion Jones, Aidan~N Gomez, \L~ukasz Kaiser, and Illia Polosukhin.
\newblock Attention is all you need.
\newblock In I.~Guyon, U.~Von Luxburg, S.~Bengio, H.~Wallach, R.~Fergus, S.~Vishwanathan, and R.~Garnett (eds.), \emph{Advances in Neural Information Processing Systems}, volume~30. Curran Associates, Inc., 2017.
\newblock URL \url{https://proceedings.neurips.cc/paper_files/paper/2017/file/3f5ee243547dee91fbd053c1c4a845aa-Paper.pdf}.

\bibitem[Vignac et~al.(2022)Vignac, Krawczuk, Siraudin, Wang, Cevher, and Frossard]{vignac2022digress}
Clement Vignac, Igor Krawczuk, Antoine Siraudin, Bohan Wang, Volkan Cevher, and Pascal Frossard.
\newblock Digress: Discrete denoising diffusion for graph generation.
\newblock \emph{arXiv preprint arXiv:2209.14734}, 2022.

\bibitem[Watts \& Strogatz(1998)Watts and Strogatz]{watts1998collective}
Duncan~J Watts and Steven~H Strogatz.
\newblock Collective dynamics of ‘small-world’networks.
\newblock \emph{nature}, 393\penalty0 (6684):\penalty0 440--442, 1998.

\bibitem[Yang et~al.(2024)Yang, Huang, Zhang, Liu, Hong, Zhang, Yang, Cui, and Zhang]{yang2024graphusion}
Ling Yang, Zhilin Huang, Zhilong Zhang, Zhongyi Liu, Shenda Hong, Wentao Zhang, Wenming Yang, Bin Cui, and Luxia Zhang.
\newblock Graphusion: Latent diffusion for graph generation.
\newblock \emph{IEEE Transactions on Knowledge and Data Engineering}, 2024.

\bibitem[Yoon et~al.(2023)Yoon, Wu, Palowitch, Perozzi, and Salakhutdinov]{yoon2023graph}
Minji Yoon, Yue Wu, John Palowitch, Bryan Perozzi, and Russ Salakhutdinov.
\newblock Graph generative model for benchmarking graph neural networks.
\newblock 2023.

\bibitem[You et~al.(2018{\natexlab{a}})You, Liu, Ying, Pande, and Leskovec]{you2018graph}
Jiaxuan You, Bowen Liu, Zhitao Ying, Vijay Pande, and Jure Leskovec.
\newblock Graph convolutional policy network for goal-directed molecular graph generation.
\newblock \emph{Advances in neural information processing systems}, 31, 2018{\natexlab{a}}.

\bibitem[You et~al.(2018{\natexlab{b}})You, Ying, Ren, Hamilton, and Leskovec]{you2018graphrnn}
Jiaxuan You, Rex Ying, Xiang Ren, William Hamilton, and Jure Leskovec.
\newblock Graphrnn: Generating realistic graphs with deep auto-regressive models.
\newblock In \emph{International conference on machine learning}, pp.\  5708--5717. PMLR, 2018{\natexlab{b}}.

\bibitem[Zaremba et~al.(2014)Zaremba, Sutskever, and Vinyals]{zaremba2014recurrent}
Wojciech Zaremba, Ilya Sutskever, and Oriol Vinyals.
\newblock Recurrent neural network regularization.
\newblock \emph{arXiv preprint arXiv:1409.2329}, 2014.

\bibitem[Zhou et~al.(2024)Zhou, Wang, and Zhang]{zhou2024unifying}
Cai Zhou, Xiyuan Wang, and Muhan Zhang.
\newblock Unifying generation and prediction on graphs with latent graph diffusion.
\newblock In \emph{The Thirty-eighth Annual Conference on Neural Information Processing Systems}, 2024.

\end{thebibliography}
\bibliographystyle{iclr2025_conference}

\newpage
\appendix

\section{Datasets.}\label{sec:dataset}
 
\begin{figure}[t]
\centering
\includegraphics[width=0.14\textwidth]{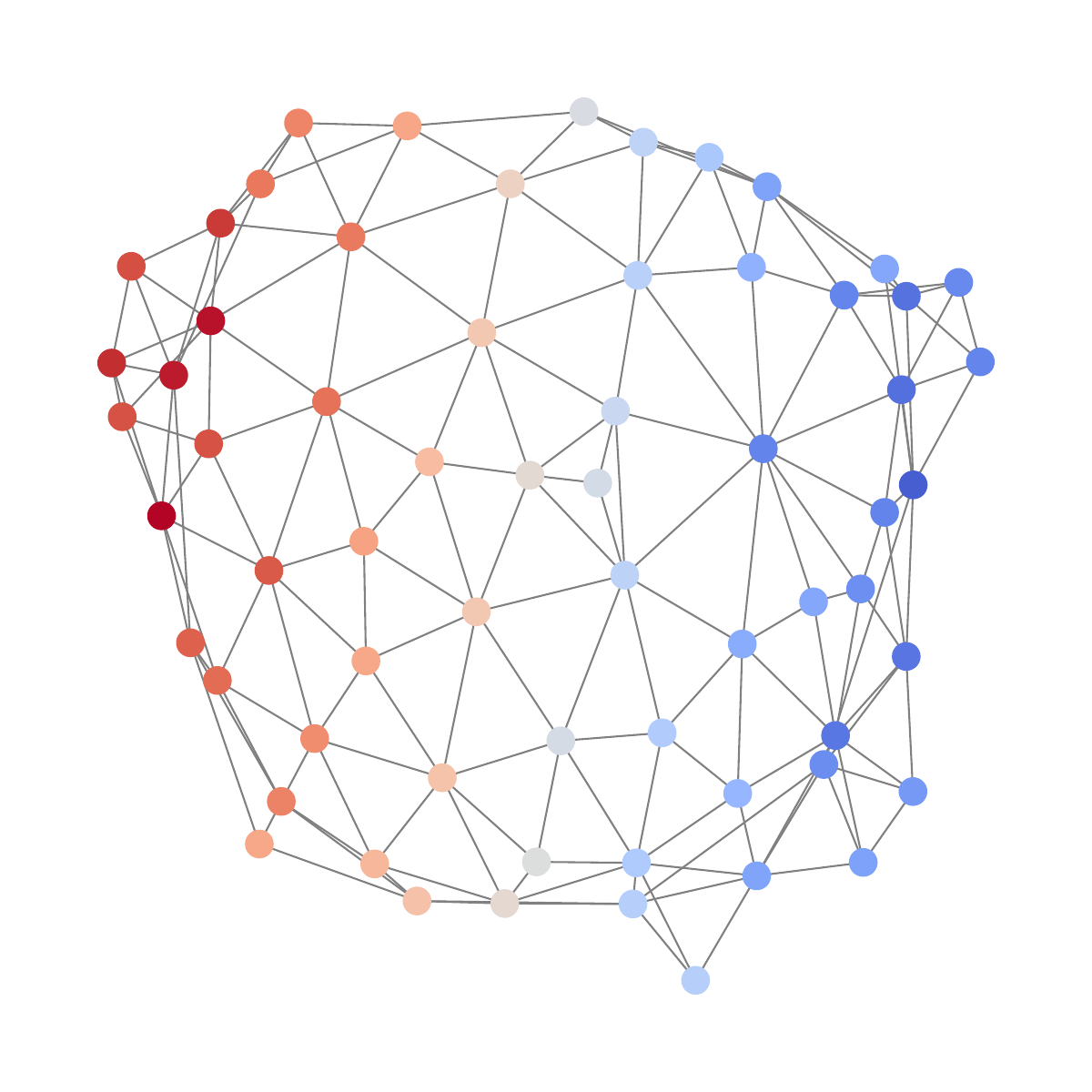} 
\includegraphics[width=0.14\textwidth]{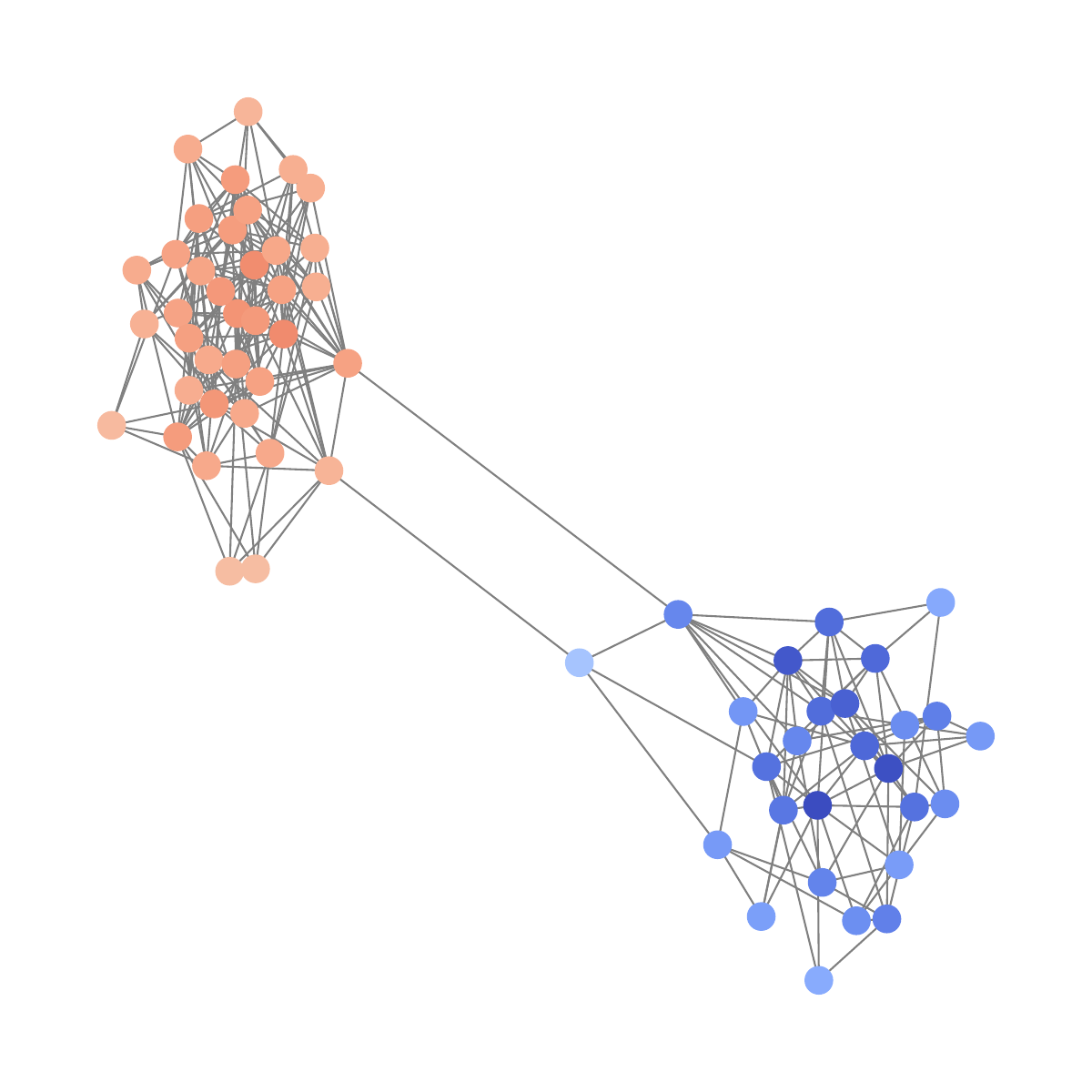} 
\includegraphics[width=0.14\textwidth]{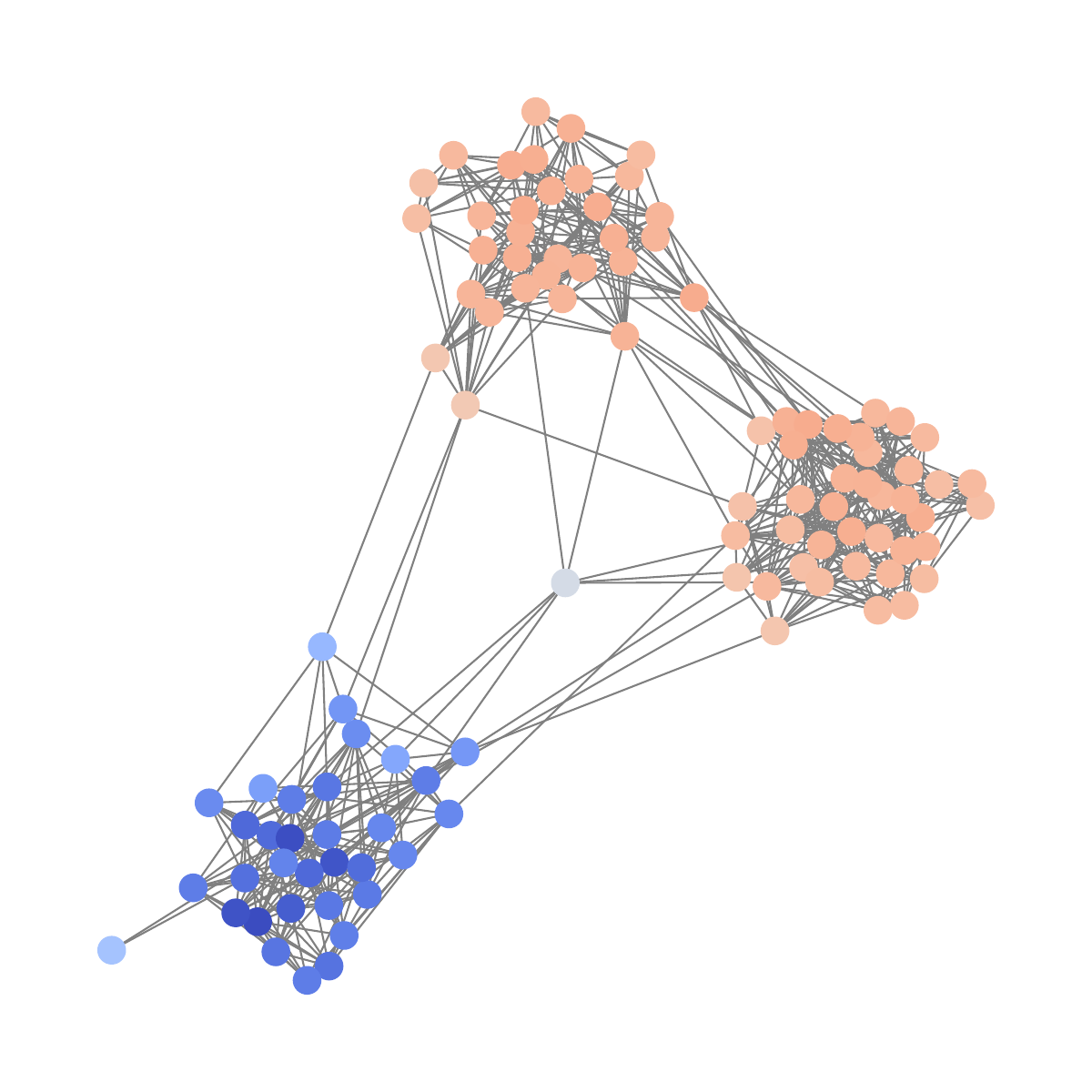}
\includegraphics[width=0.14\textwidth]{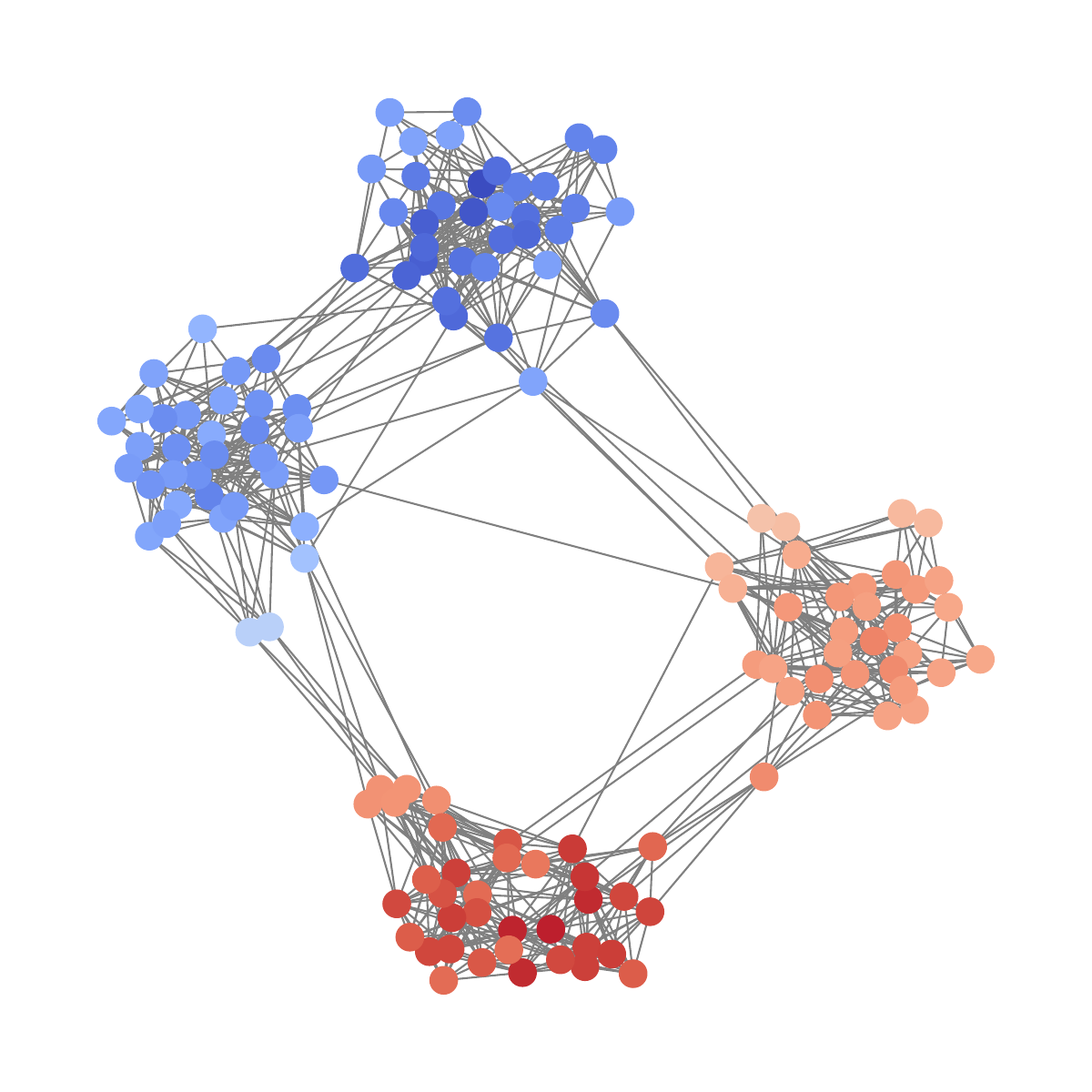}   
\includegraphics[width=0.14\textwidth]{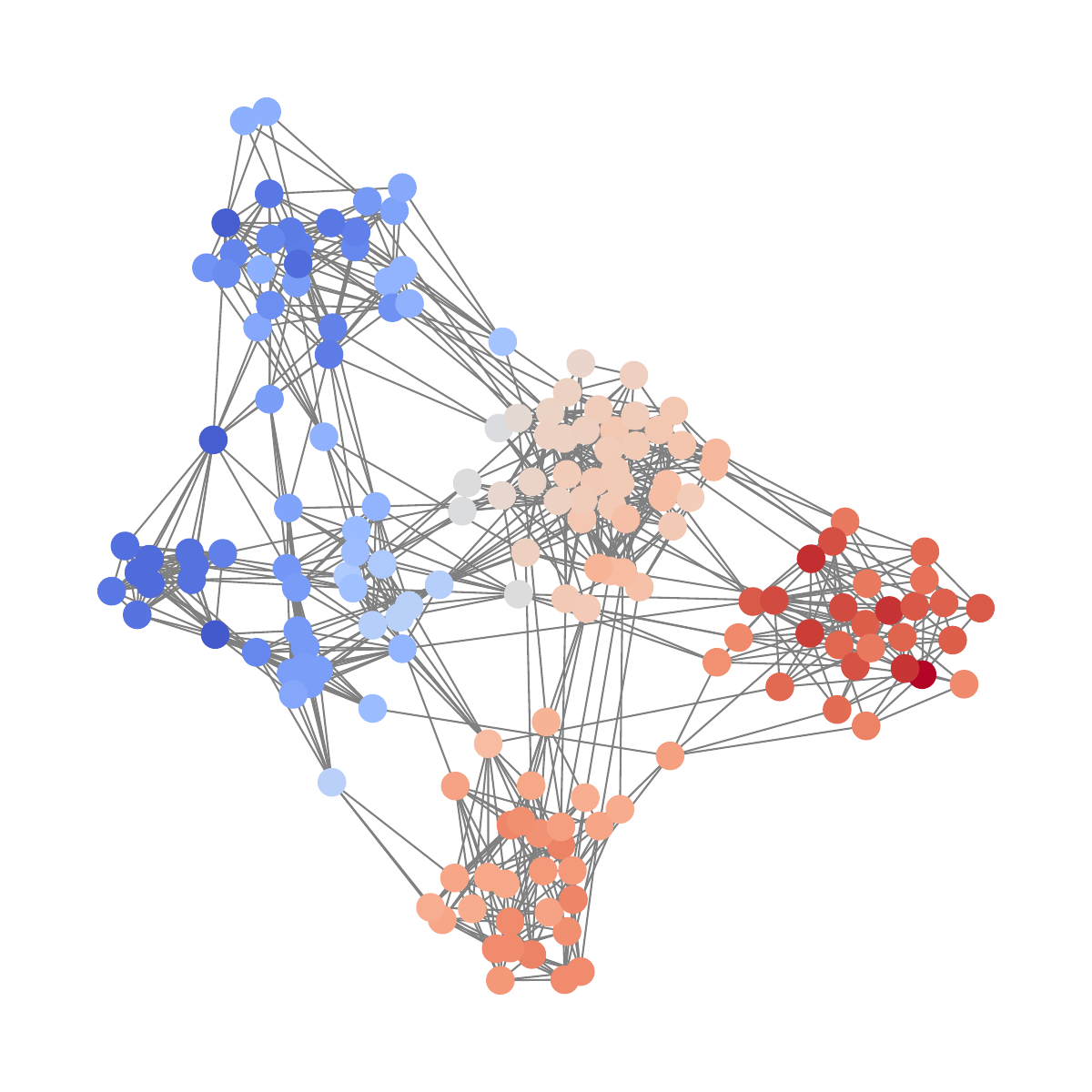}
\includegraphics[width=0.14\textwidth]{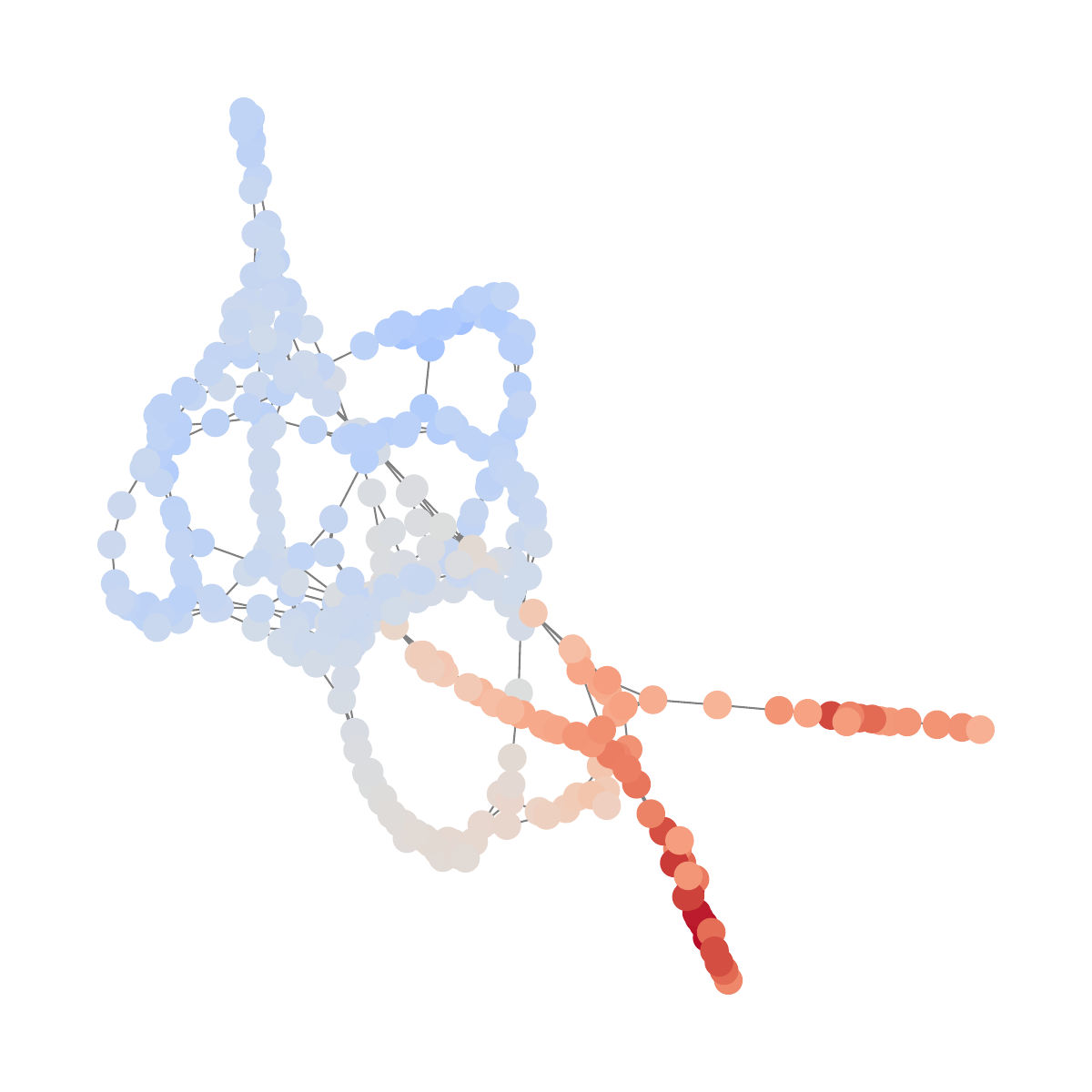}

\copyrightbox[b]{
\includegraphics[width=0.15\textwidth]{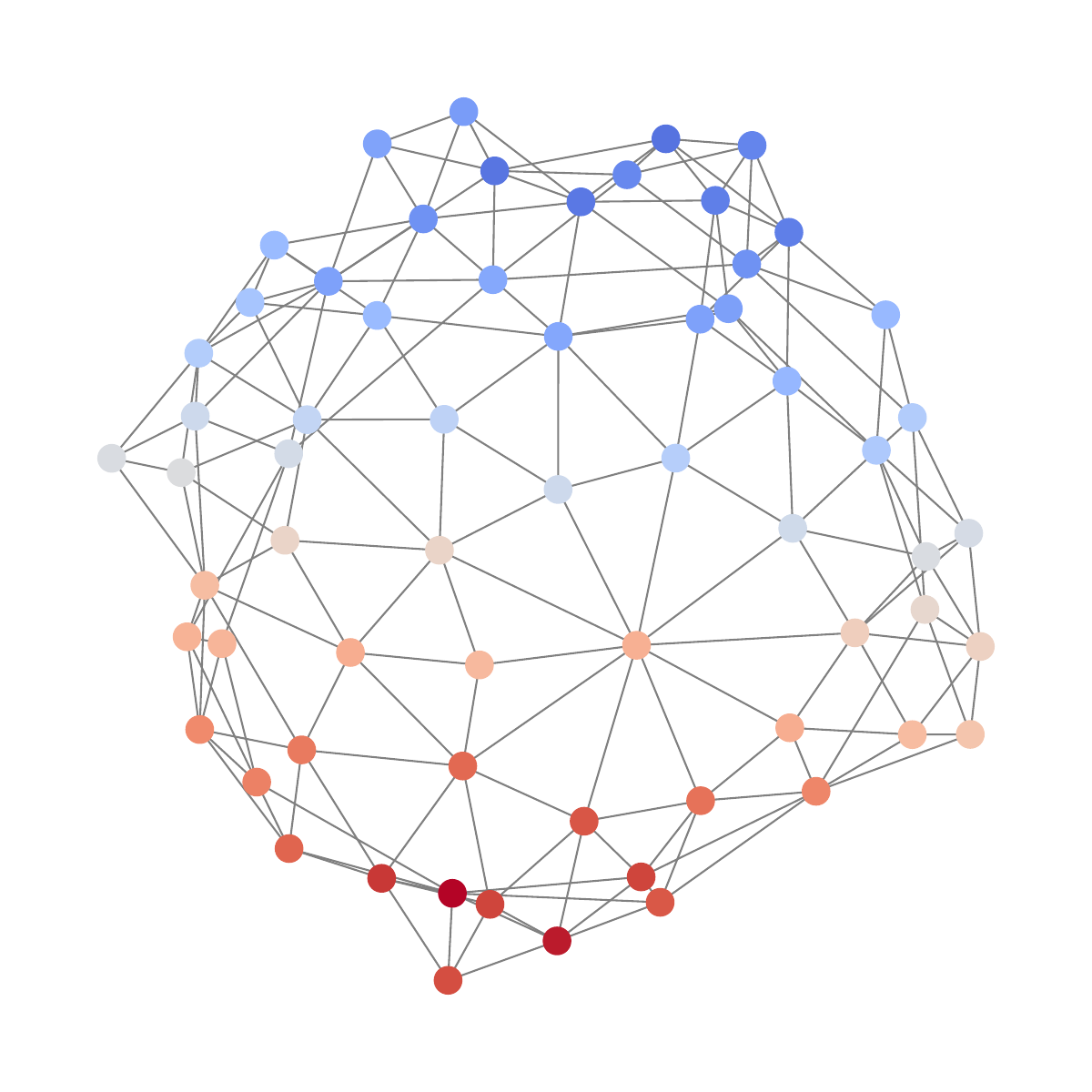} }{\hspace{25pt}Planar}
\copyrightbox[b]{\includegraphics[width=0.14\textwidth]{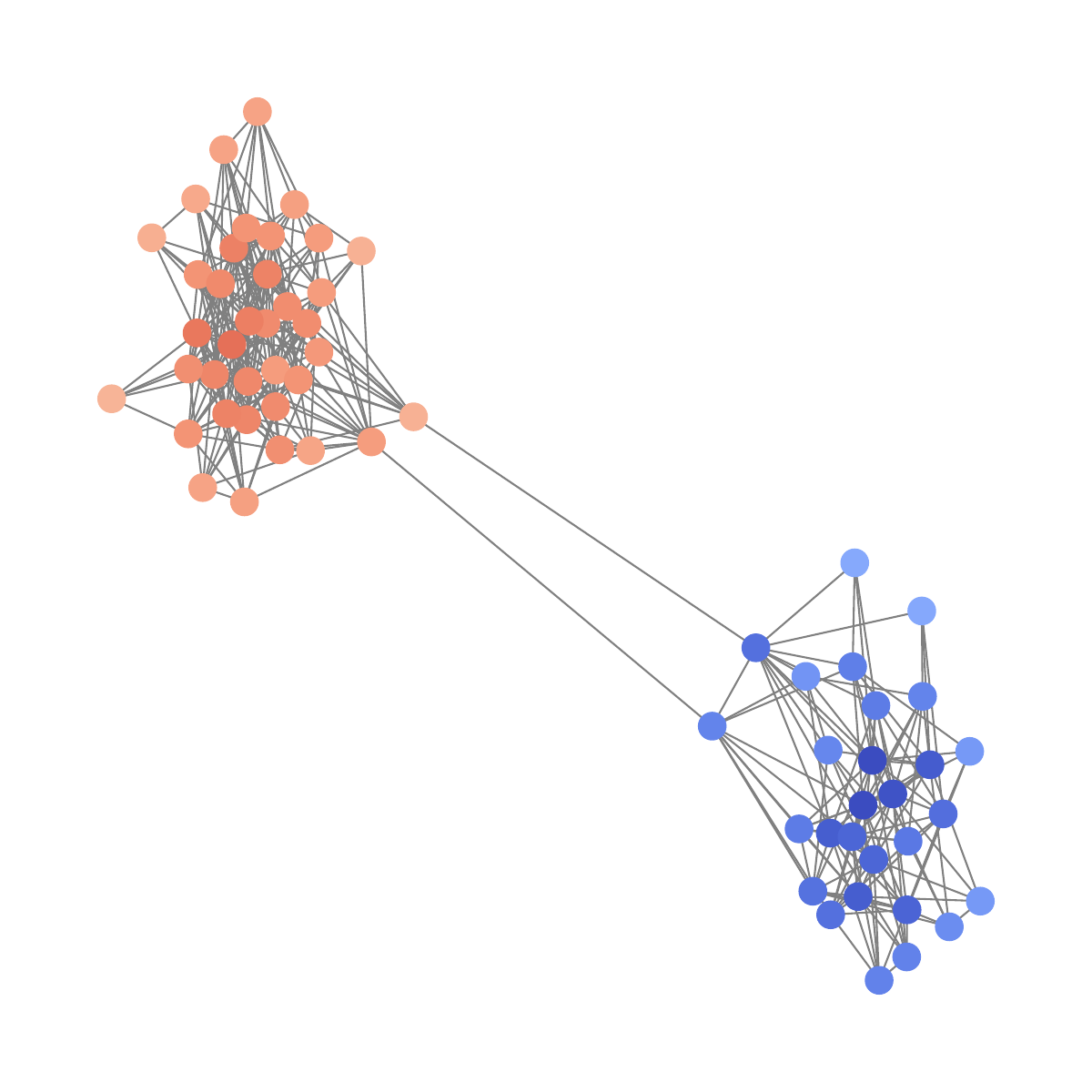}}{\hspace{100pt}SBM 2}
\copyrightbox[b]{  
\includegraphics[width=0.14\textwidth]{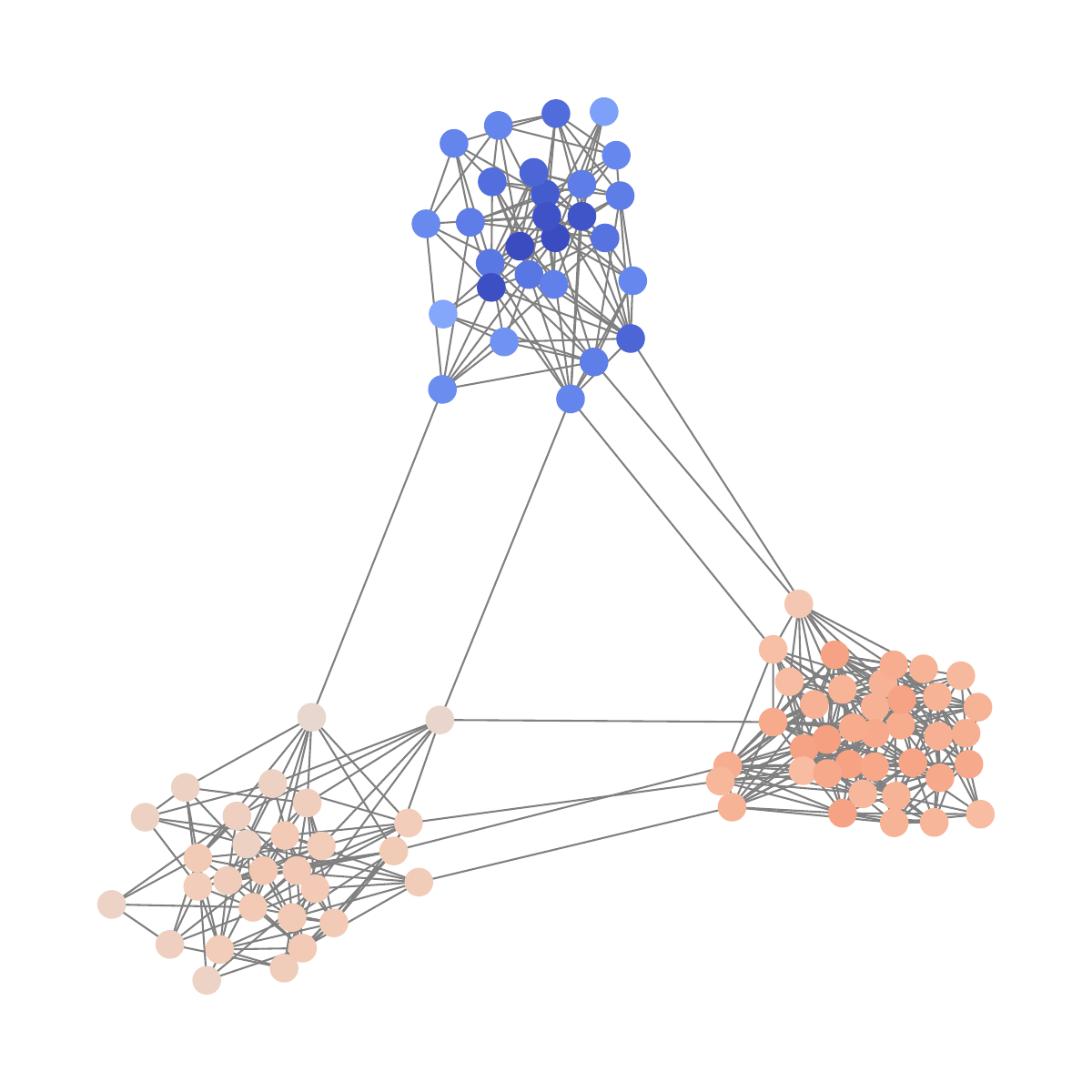} }{\hspace{50pt}  SBM 3}
\copyrightbox[b]{   
\includegraphics[width=0.14\textwidth]{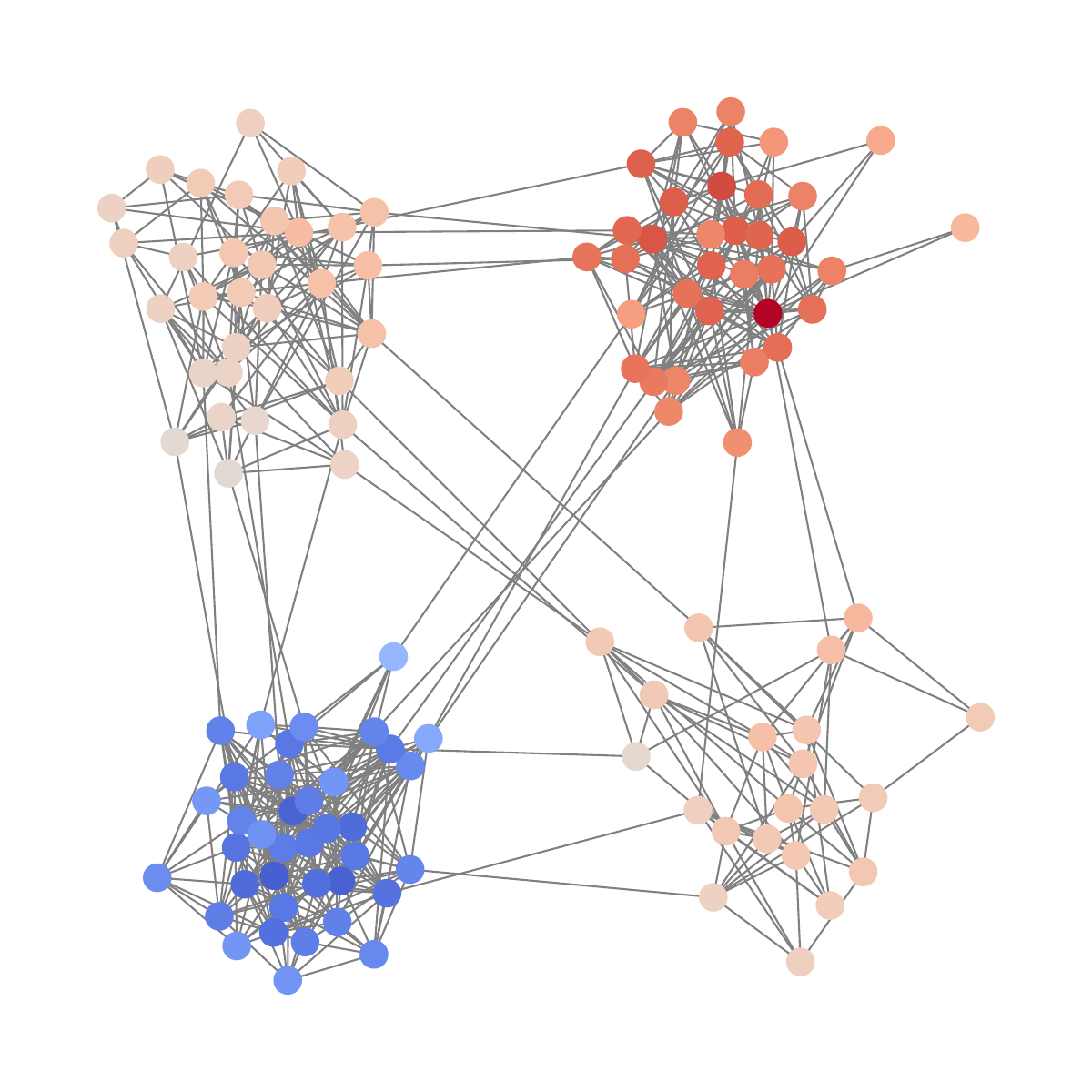}  }{\hspace{50pt}  SBM 4}
\copyrightbox[b]{\includegraphics[width=0.14\textwidth]{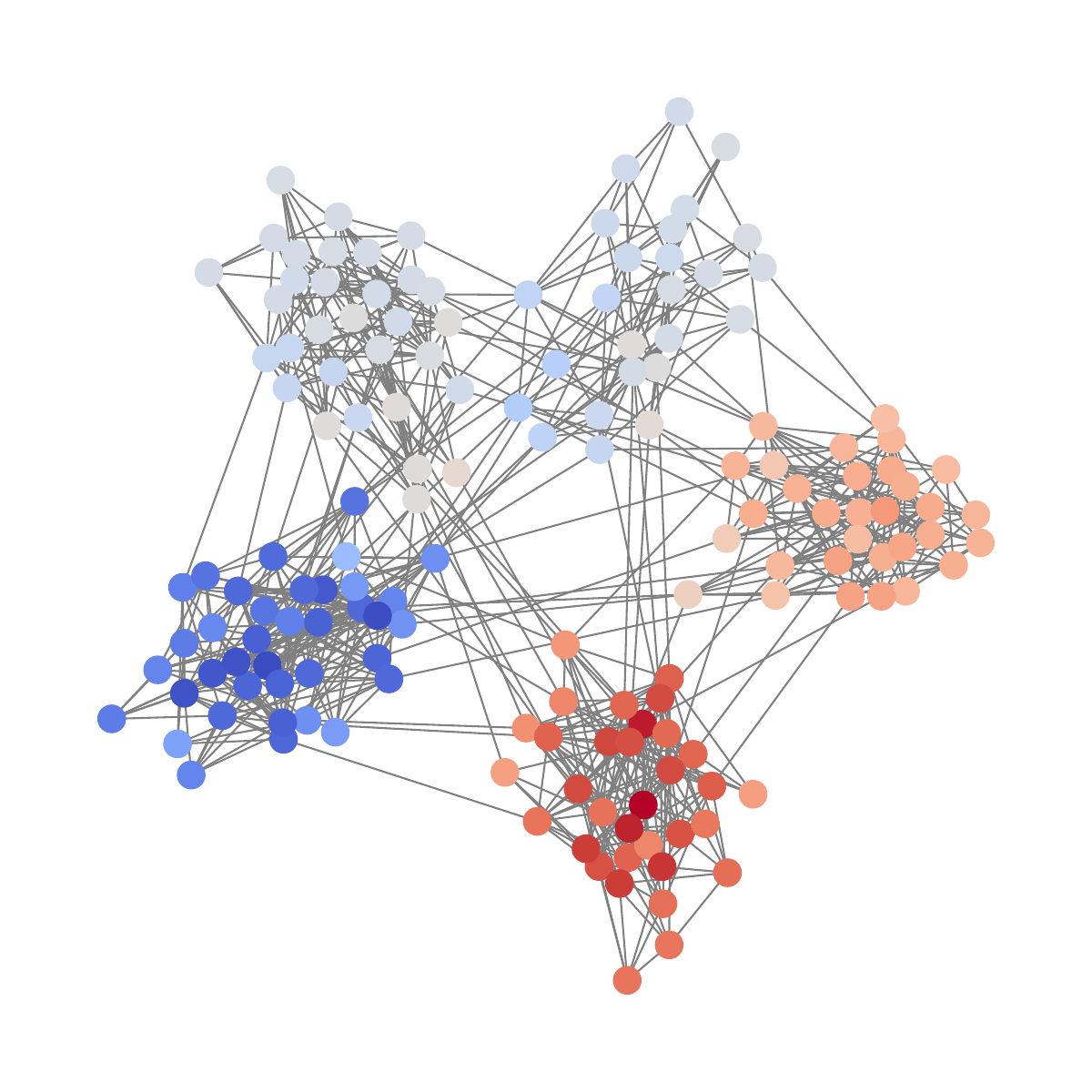}  }{\hspace{50pt}  SBM 5}
\copyrightbox[b]{
\includegraphics[width=0.14\textwidth]{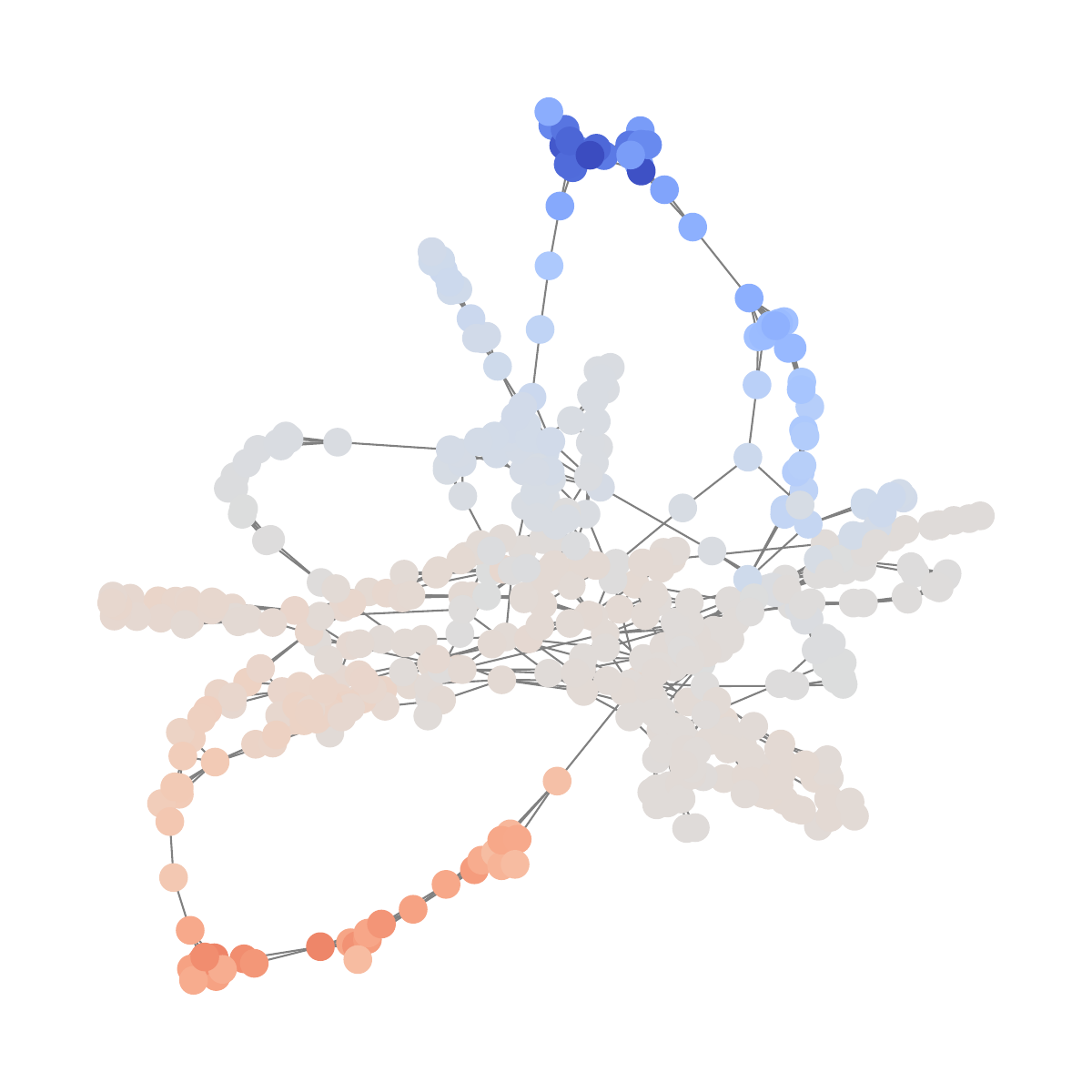}}{Proteins}

\caption{Visual comparison between training set graph samples and generated graph samples produced by \grasp. Each column represents a graph type (Planar, Stochastic Block Model with 2,3,4, and 5 communities, and Proteins). Top row  (Original): training set graphs. Bottom row (Generated): graph generated by \grasp.}
\label{fig:qualitative}
\end{figure}

We utilize five commonly used datasets for graph generative tasks. Some examples of the graphs contained in these datasets alongside a similar graph generated by \grasp~are shown in \Cref{fig:qualitative}.

\textbf{Community-small}: A synthetic dataset consisting of 100 random community graphs, 
with $[12,20]$ nodes.

\textbf{Stochastic Block Model (SBM)}: A synthetic  dataset from \cite{martinkus2022spectre} comprising 200 Stochastic Block Model graphs, each with a random selection of 2 to 5 communities and 20 to 40 nodes within each community. The probability of edges between communities is set at $0.3$, while the probability of edges within communities is set at $0.05$. 

\textbf{Planar}: A synthetic dataset from \cite{martinkus2022spectre} of 200 planar graphs, each containing 64 nodes. Graphs are created through the application of Delaunay triangulation to a randomly and uniformly placed set of points.

\textbf{QM9} \cite{ramakrishnan2014quantum, ruddigkeit2012enumeration}: 
This real-world dataset comprises 134k organic molecules with a maximum of 9 heavy atoms (carbon, oxygen, nitrogen, and fluorine). By following \cite{simonovsky2018graphvae} we allocate 10k molecules for validation, 10k for testing, and the rest for training.   
 
\textbf{Proteins} \cite{dobson2003distinguishing}:  The dataset encompasses 918 protein graphs, each ranging from 100 to 500 nodes. In this representation, each protein is depicted as a graph, with nodes corresponding to amino acids. Nodes are connected by an edge if they are within a distance of less than 6 Angstroms from each other.

\section{Evaluation Metrics}\label{sec:metrics}
\subsection{Statistics-Based}
We consider the following key graph statistics: degree distribution (Deg.), clustering coefficient (Clus.), and the occurrence frequency of all 4-node orbits (Orb.). The deviation of these metrics between the generated graphs and the actual ones is measured using the maximum mean discrepancy (MMD)~\cite{you2018graphrnn}. In its initial formulation, the computation of the MMD relied on the Earth Mover's Distance (EMD) and as a result was very slow. For this reason, as suggested in~\cite{liao2019efficient}, we use the total variational (TV) Gaussian kernel. This in turn significantly accelerates the evaluation process while maintaining consistency with EMD. In addition to assessing node degree, clustering coefficient, and orbit counts, we also extend our evaluation to include a spectral analysis (Spect.), following~\cite{liao2019efficient}. This involves computing the eigenvalues of the normalized graph Laplacian, quantized to approximate a probability density. The spectral comparison offers insights into the global properties of the graphs, complementing the local graph statistics emphasized by previous metrics.
\subsection{Intrinsic-Quality-Based}
The \emph{validity} is determined by the ratio of valid molecules to all generated molecules. For molecule graphs (QM9), the validity in molecule generation represents the percentage of chemically valid molecules based on specific domain rules. We measure it using RDKit anitization\footnote{\url{https://www.rdkit.org/docs/RDKit\_Book.html}}. \revised{For Planar graphs, we use the NetworkX python library based on the left-right planarity test \citep{DEFRAYSSEIX2012279}. For the stochastic block model graphs (SBM), we use a Monte Carlo and greedy heuristic for the inference of the stochastic block model parameters (as implemented by cdlib \citep{sbmgreedy}). We consider a graph valid if its probability, assessed by a Wald test \citep{fahrmeir2013regression}, of having been generated by the SBM model with inter-edge probability of 0.3 and intra-edge probability of 0.005 is higher than 0.9, and if the number of communities is in the range 2-5 with a number of nodes for each community between 20 and 40.
}

The \emph{novelty} gauges the percentage of \revised{valid} graphs that are not sub-graphs of the training set, and vice versa. It checks if the model has successfully learned to generalize to unseen graphs and it considers two graphs identical if they are isomorphic.

The \emph{uniqueness} is defined as the ratio of unique samples to valid \revised{and novel} samples, measuring the level of variety during sampling. To calculate uniqueness, generated graphs that are sub-graph isomorphic to others are initially removed, and the remaining percentage represents uniqueness. For instance, if a model generates 100 identical graphs, the uniqueness is $ 1/100 = 1\%$.

\revised{The product of these three metrics is referred to as V\&U\&N (or VUN) and summarizes the ability of the method to generate graphs that are at the same time novel, unique, and valid.} 

\section{\revised{VUN Experiments}}\label{sec:vun}

\begin{table}
\setlength{\tabcolsep}{0.65\tabcolsep}
\caption{\revised{Comparison with other graph generative models  based on validity, uniqueness, and novelty metrics (the higher the better) on synthetic datasets.
\label{tab:vun_synth}}}

\scriptsize
\centering 
 
\begin{tabular}{@{}rcccc|cccc@{}}
\toprule   
 & \multicolumn{4}{c}{\textbf{Planar} }  & \multicolumn{4}{c}{\textbf{Stochastic Block Model (SBM)} }      \\ 
 
\cmidrule{2-9}  
& Val.$\uparrow$&  Uniq.$\uparrow$&   Nov.$\uparrow$ & V\&U\&N $\uparrow$ &   Val.$\uparrow$&  Uniq.$\uparrow$&   Nov.$\uparrow$ & V\&U\&N $\uparrow$  \\
\midrule 
\midrule 
GraphRNN & 0.00 & -    & -    & 0.00 & 0.13 & 1.00 & 1.00 & 0.13 \\
GRAN     & 0.03 & 1.00 & 1.00 & 0.03 & 0.20 & 1.00 & 1.00 & 0.20 \\
DiGress  & 0.70 & 0.98 & 0.95 & \textbf{0.65} & 0.13 & 0.98 & 1.00 & 0.13 \\
GSDM     & 0.85 & 0.50 & 0.28 & 0.12 & 0.08 & 0.88 & 0.50 & 0.04 \\
GDSS     & 0.00 & -    & -    & 0.00 & 0.01 & 1.00 & 1.00 & 0.01 \\
SPECTRE  & 0.14 & 1.00 & 1.00 & 0.14 & 0.51 & 1.00 & 1.00 & \textbf{0.51} \\
 
\midrule
\grasp$\;\,$ & 0.15 & 1.00 & 1.00 & \underline{0.15} &  0.49 & 1.00 & 1.00 & \underline{0.49} \\
\bottomrule 
\end{tabular} 
\end{table}
\revised{ \Cref{tab:vun_synth} presents the results on the quality of the synthetic graphs generated for the Planar and SBM datasets, analyzing the metrics of validity, uniqueness, novelty, and their combination. 
As highlighted in \Cref{par:synth_res}, our method achieves 100\% uniqueness and novelty on both datasets while showing the second-best score in terms of VUN. Specifically, for SBM, the gap of SPECTRE and our method with respect to other methods is considerable, due to a higher validity score compared to competitors. On the other hand, the validity score of our method on the Planar dataset is lower than the top-performing method (DiGress) of some margin. This behavior is not unexpected and can be explained by the insensibility of the eigenvectors to small local changes of the topology as discussed in \Cref{par:synth_res}.
It is worth noticing how GSDM, despite showing the best validity performance on planar graphs, is not able to generate novel graphs, with a novelty score of just 28\% in Planar, and 50\% in SBM. This does not come as a surprise, since in GSDM the eigenvectors used to reconstruct the final adjacency matrix are uniformly sampled from the training set. This limits the generative power of the method, since most of the information about the graph connectivity is contained in the eigenvectors. As such, the obtained graphs are not actually generated but rather slight modifications of the training set graphs.}

\section{Model Settings and Implementation Details}\label{sec:modelsettings}
For all datasets except for QM9, we retrained the models using the configurations recommended by the authors.
When no recommended hyper-parameters setups or model weights were available, we explored the space of hyper-parameters tuning them according to the ranges mentioned for other datasets. Finally, for our method (\grasp), we use the $k$ largest/smallest eigenvalues of the unnormalized Laplacian.
The values of $k$ are experimentally determined as explained in Appendix~\ref{sec:num_eigenvectors}. We stress that these are only a fraction of the full set of eigenvectors.

\red{We used the unnormalized Laplacian since it yields an easier graph reconstruction by simple thresholding (as explained  in \Cref{sec:ablation}). In order to handle potential scaling issues, we simply normalized the eigenvalues and eigenvectors based on the training data so as to reflect a normal distribution.}

For the training of the diffusion model, we split each dataset into 90\% train and 10\% test, and we train the Spectral Diffusion on the whole dataset for 100k epochs, using early stopping on the reconstruction loss. We performed a grid search on the number of layers between 6, 9 and 12, and selected the best model according to the degree metric computed from the graphs reconstructed directly from the eigenvectors/values and the graphs of the training set. The sampling has been done using DDIM with 200 steps. Moreover, we generate each sample 4 times and keep the one with the lower deviation from orthogonality.

For the training of the Graph Predictor, we used the same splits of the Spectral Diffusion, and trained for 100k epochs. We performed early stopping by comparing the degree distribution of the generated graphs with the training graphs. We used 6 PPGN layers and 3 PPGN layers for the Graph Predictor and the discriminator network respectively, except for QM9 in which also the Graph Predictor is composed of three layers. For QM9, we let the Graph Predictor to generate also edge features, similarly to \cite{martinkus2022spectre}. 

For all datasets, following the observations in Appendix~\ref{sec:num_eigenvectors}, we train both Spectral Diffusion and Predictor on the 16 smallest and 32 largest eigenpairs and select the final model according to the best average metrics on the validation set.

In order to guarantee the reproducibility of both our model architecture and results, we have made our code accessible on an online public repository
\footnote{ \url{https://github.com/lcosmo/GGSD}}.

\begin{figure*}[t!]
\centering
\includegraphics[width=0.463\textwidth]{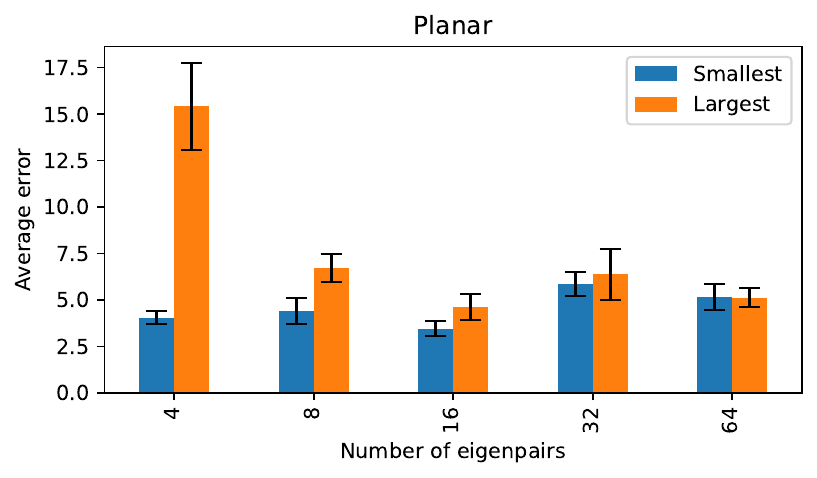}
\includegraphics[width=0.45\textwidth]{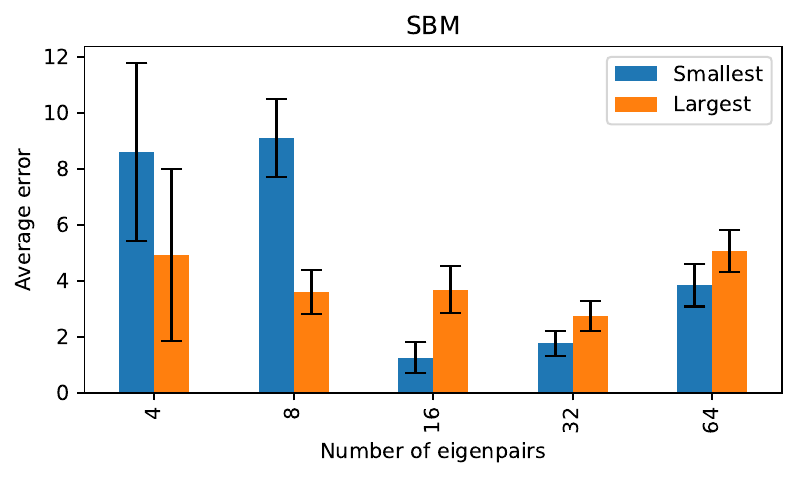}
\caption{
Performance analysis with varying numbers of eigenpairs and different spectrum parts, for the SBM (left) and Planar (right) datasets. The average error represents the mean degradation of all metrics between the generated graphs and the training set.  We report both the mean and the standard deviation as error bars on 10 generations of 200 graphs. Specifically, we consider the Degree, Cluster, and Spectral metrics. \revised{For each metric, the degradation is computed as the ratio between (1) the MMD value computed between the generated graphs and the training graphs and (2) the MMD computed between the test graphs and the training graphs. A value of 1 indicates that the generated graphs exhibit the same statistical difference wrt the training graphs as the test set graphs do. The ratios computed for each metric are then averaged to get a single value indicating the quality of the generation.}}
\label{fig:numeig}  
\end{figure*}

\section{Number of Eigenvectors}\label{sec:num_eigenvectors}
To evaluate which part of the spectrum is more relevant and the proper number of eigenvectors to use, we performed an experiment.  We trained our model focusing on either the smallest eigenvalues (Smallest) or the largest ones (Largest), while gradually increasing the corresponding number of eigenvectors taken into consideration. The results obtained from two synthetic datasets, Planar and SBM,  are shown in \Cref{fig:numeig}. Regarding the optimal number of eigenvectors, it appears that too many eigenvectors do not yield the best results, either considering low or high frequencies. Specifically, in the case of the Planar dataset, both smaller and larger eigenvalues exhibit the best performance with 16 eigenvectors. However, for SBM, while the optimal count for lower frequencies remains at 16, for higher frequencies, it increases to 32. These results are not entirely surprising, considering the inherent trade-off between the diffusion model's capability to manage high-dimensional data and the quantity of information (number of eigenvectors) accessible to the Predictor. All in all, it should be noted that the selection of the number of eigenvectors can generally be regarded as a model hyperparameter for optimization, acknowledging its potential dependence on the specific dataset.

Building on the findings outlined above, in our experiments we employed the 32 largest eigenvalues and the 16 smallest eigenvalues - and their corresponding eigenvectors - for all datasets, except for the community-small dataset, for which we utilized the top  8 largest/smallest eigenvalues, and QM9, for which we used the full set of eigenpairs.

\section{Eigenvectors Orthogonality Study}\label{sec:apportho}

\begin{figure}[b]
    \centering
    \begin{overpic}[trim=0mm -3mm 0 0mm, clip, width=0.315\textwidth]{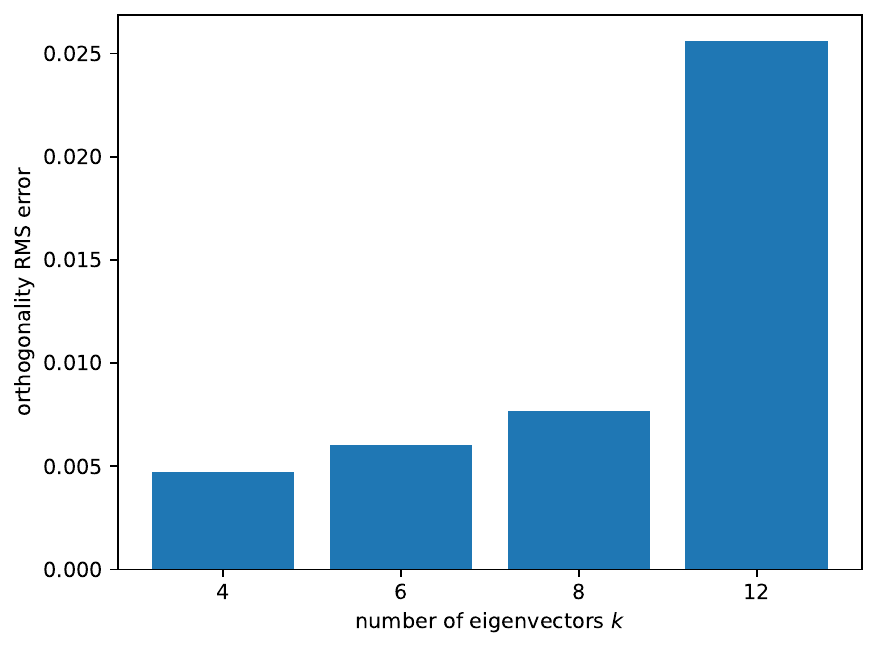}
        \put(40,78){\scriptsize{Community}}
    \end{overpic}
    \begin{overpic}[trim=0mm -3mm 0 0mm, clip, width=0.315\textwidth]{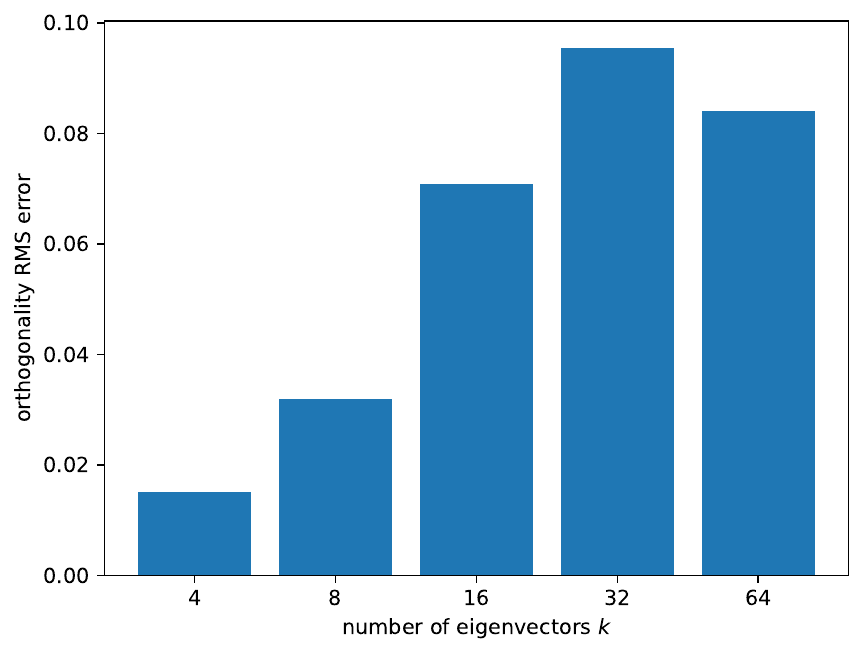}
        \put(45,78){\scriptsize{SBM}}
    \end{overpic}
    \hfill
    \begin{overpic}[trim=0mm -0mm 0 -20mm, clip, width=0.325\textwidth]{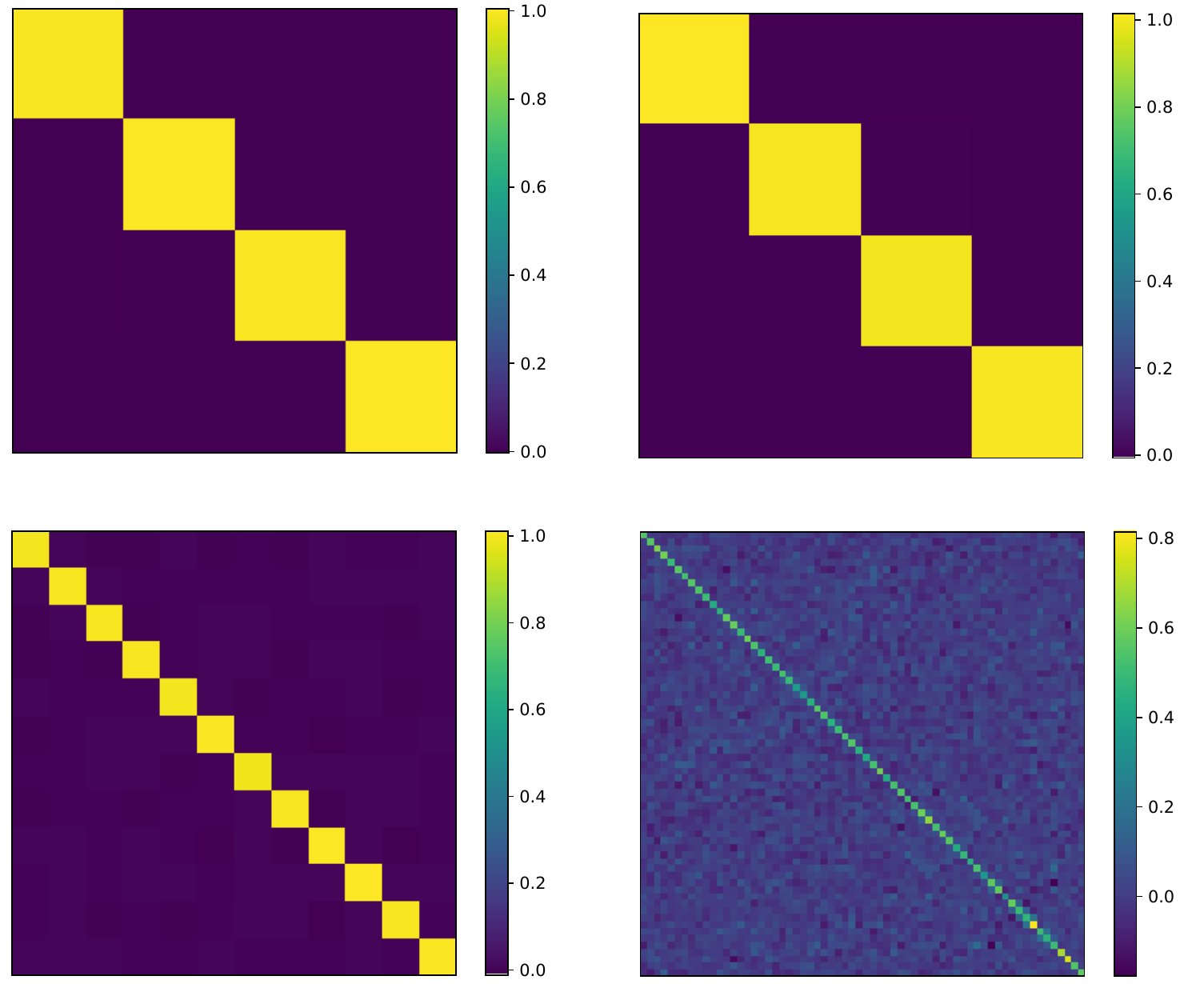}
        \put(7,88){\scriptsize{Community}}
        \put(67,88){\scriptsize{SBM}}
        \put(66,82){\scriptsize{$k=4$}}
        \put(64,38.5){\scriptsize{$k=64$}}
        \put(12,82){\scriptsize{$k=4$}}
        \put(10,38.5){\scriptsize{$k=12$}}
    \end{overpic}
    \caption{\red{\textbf{Left}: the two bar plots show the deviation of the eigenvectors generated by \grasp~from an orthonormal basis on two datasets for different numbers of eigenpairs. The deviation from orthogonality is computed as the average root mean squared difference of the inner product of the eigenvectors with the identity matrix, {\em i.e.}, $RMS (\mathbf{\tilde{\Phi}_0}) = (\frac{1}{k^2} \mathbf{1}(\mathbf{\tilde{\Phi}_0}^\top \mathbf{\tilde{\Phi}_0} - \mathbf{I})^{\cdot 2} \mathbf{1}^\top)^{\frac{1}{2}}$, with $^{\cdot 2}$ being the elementwise square operator and $k$ the number of eigenvectors. \textbf{Right}: Qualitative results showing the inner product of the generated eigenvectors. \label{fig:err_vs_k}}}
\end{figure}

\red{We conducted some experiments to provide both quantitative and qualitative analyses of the orthogonality behavior of the generated eigenvectors. In \Cref{fig:err_vs_k} (left), we show how the generated eigenvectors deviate from forming an orthonormal basis. \revised{Here we vary the number of generated eigenvectors between 4 and 12 for Community, and between 4 and 64 for SBM. This choice reflects the fact that the graphs in the Community dataset have between 12 and 20 nodes, while in the case of SBM we observed that using more than 64 eigenvectors appears to lead to a degradation in performance (see Figure~\ref{fig:err_vs_k}).}
As expected, increasing the number of generated eigenvectors introduces greater deviations from orthogonality, which aligns with our findings in  \Cref{sec:num_eigenvectors}. In \Cref{fig:err_vs_k} (right), we provide qualitative examples by comparing the generated eigenvectors with those computed from the adjacency matrix predicted by the PPGN.   We can observe that in simpler datasets, such as the Community dataset, the eigenvectors are perfectly aligned. In more challenging datasets, such as SBM, while the alignment is not exact, the overall correspondence remains good and significant.}

\red{The two experiments described above demonstrate that the generated eigenvectors exhibit the smoothness property and are often very similar, sometimes nearly identical, to the eigenvectors computed from the final predicted graph. To further investigate this, we computed the Dirichlet energy of the generated approximate eigenvectors on the final generated graph. We observed a consistent pattern, with the energy increasing as the eigenvalues grow larger. In \Cref{tab:Dirichlet}, we report the average Dirichlet energy computed from 500 generated graphs in the SBM and Community datasets. We exclude the first eigenvector ($\lambda=0$) from the analysis, as it does not contribute to the reconstruction of the Laplacian, and the diffusion process is trained only on non-zero eigenvalues.}

\begin{table}[t!]
\centering
\scriptsize
\renewcommand{\arraystretch}{1.2}
\setlength{\tabcolsep}{3pt}
\resizebox{\textwidth}{!}{
\begin{tabular}{c cccccccccccccccc}
\hline
{$\mathbf{k}$} & 1 & 2 & 3 & 4 & 5 & 6 & 7 & 8 & 9 & 10 & 11 & 12 & 13 & 14 & 15 & 16 \\ \hline \hline
{SBM} & 1.11 & 2.54 & 3.62 & 4.88 & 6.65 & 6.88 & 7.11 & 7.42 & 7.51 & 7.75 & 7.47 & 7.84 & 7.76 & 7.85 & 8.26 & 8.13 \\ 
{Community} & 1.09 & 2.97 & 3.41 & 4.18 & 4.71 & 5.46 & 5.77 & 6.13 & 6.46 & 6.79 & 6.87 & 7.54 & - & - & - & - \\ \hline
\end{tabular}
}
\caption{Average Dirichlet energy computed on 500 generated graphs from the SBM and Community datasets, trained on the lower part of the spectrum (smaller eigenvalues).}
\label{tab:Dirichlet}
\end{table}

\section{Runtime Comparison} \label{sec:runtime}
\begin{table}[h]
\setlength{\tabcolsep}{0.7\tabcolsep}
\caption{For each dataset, every method generates 100 graphs. We report the total generation time in seconds.} 
\label{tab:runtime}
\centering  
\begin{tabular}{@{}rrrrrr@{}}
\toprule   & \textbf{Planar} & \textbf{SBM} & \textbf{Comm.} & \textbf{Proteins} & \textbf{QM9} \\ 
\midrule 
\midrule 
GraphRNN & \underline{3.30} & \underline{4.19} & 3.16 & \textbf{16.87} & --- \\
GRAN & 8.58 & 48.43 & \underline{2.90} & 205.47 & --- \\
DiGress & 859.64 & 3882.30 & 70.49 & OOM & 68.11 \\
\revised{GSDM} & 10.71 & 31.51 & 9.74 & 160.09 & --- \\
GDSS & 81.10 & 160.88 & 456.30 & 3177.03 & \underline{1.32} \\
SPECTRE & \textbf{1.71} & \textbf{3.63} & \textbf{0.72} & OOM & \textbf{0.06} \\
\midrule
\grasp & 9.51 & 18.63 & 4.41 & \underline{124.07} & 2.50 \\
\bottomrule 
\end{tabular} 
\end{table}

Table~\ref{tab:runtime} shows the total generation time (in seconds) for our method and the baseline methods across the datasets analyzed in this paper. For a fair comparison, we generated 100 graphs for each dataset and method. These experiments were conducted on a computer equipped with an AMD Ryzen 7 3700X processor, 64GB of RAM, and an NVIDIA RTX 3070 8GB graphics card. We achieve a significant speedup compared to DiGress, thus showing that we are able to overcome the computational bottleneck of diffusion-based methods which, unlike \grasp, work on a diffusion space that is quadratic in the number of nodes of the graph. Additionally, our hardware configuration could not register the time for large datasets like Proteins due to their space complexity. While our method outperforms GDSS in terms of speed, it is less efficient than simpler algorithms such as GRAN and GraphRNN. 

\revised{\section{Stability of the Spectral Decomposition of the Graph  Laplacian}}\label{sec:instab}

\revised{In the spectral graph theory literature, the instability of the Laplacian has become a true-ism. Yet, this claim requires further qualification as several spectral approaches have shown to be robust even under severe deformation~\cite{rodola2017partial,cosmo2016matching}. In general, random structural perturbation can cause major topological changes which will reflect on the eigenvectors and eigenvalues of the Laplacian quite dramatically, but it strongly depends on the location of the actual perturbation, and it is linked with small gaps in the eigenvalues.}

\revised{From spectral perturbation theory, we note that under a perturbation $\mathscr{E}$, as long as the eigenvalues are and remain distinct, the eigenvalues of the perturbed Laplacian $\tilde{\mathbf{L}}=\mathbf{L}+\mathscr{E}$ are perturbed by a quantity
\begin{equation}
\Delta\lambda_i \approx \phi_i^T \mathscr{E} \phi_i,
\end{equation}
while the eigenvectors are perturbed by $\Delta \mathbf{\Phi} \approx \mathbf{\Phi}\mathbf{B}$. 
Here the matrix $\mathbf{B}=(b_{ij})$ is defined as:
\begin{equation}
b_{ij} = \frac{\phi_i^T\mathscr{E}\phi_j}{\lambda_j-\lambda_i}.
\end{equation}
As a consequence, the mixing can become large even for small perturbations if the gap between the eigenvalues is small, and in general only eigenvectors with close eigenvalues will mix in a significant way. All this being said, this characterizes what happens when we perturb the graph, which is not what is happening here. By recreating the spectrum through a stable diffusion process, the perturbation is in the spectrum, and, in general, small perturbations of the spectrum do not cause major topological changes in the structure (which, as we said, are associated with large spectral variations). 
Let us say that the eigenvectors are perturbed by a factor of $\Delta\mathbf{\Phi}$ and the eigenvalues by a factor of $\Delta\mathbf{\Lambda}$, then the reconstructed Laplacian is
\begin{multline}
 \tilde{\mathbf{L}} =(\mathbf{\Phi}+\Delta\mathbf{\Phi})(\mathbf{\Lambda}+\Delta\mathbf{\Lambda})(\mathbf{\Phi}+\Delta\mathbf{\Phi})^T =
\mathbf{\Phi}\mathbf{\Lambda}\mathbf{\Phi}^T + \underbrace{\mathbf{\Phi}\Delta\mathbf{\Lambda}\mathbf{\Phi}^T + \mathbf{\Phi}\mathbf{\Lambda}\Delta\mathbf{\Phi}^T + \Delta\mathbf{\Phi}\mathbf{\Lambda}\mathbf{\Phi}^T}_{\textnormal{I order error terms}} + \\
\underbrace{\mathbf{\Phi}\Delta\mathbf{\Lambda}\Delta\mathbf{\Phi}^T + \Delta\mathbf{\Phi}\Delta\mathbf{\Lambda}\mathbf{\Phi}^T + \Delta\mathbf{\Phi}\mathbf{\Lambda}\Delta\mathbf{\Phi}^T}_{\textnormal{II order error terms}} + \underbrace{\Delta\mathbf{\Phi}\Delta\mathbf{\Lambda}\Delta\mathbf{\Phi}^T}_{\textnormal{III order error term}}\,,
\end{multline}
which varies smoothly with noise and does not have elements at the denominator that force the terms to explode. Indeed, we have not observed topological instabilities in any of the datasets considered in this study.}

\end{document}